\documentclass[final,3p,times]{elsarticle}

\usepackage{amssymb}
\usepackage[fleqn]{amsmath, mathtools}
\usepackage{algorithm}
\usepackage[noend]{algpseudocode}
\usepackage{xparse}
\usepackage{setspace}
\usepackage{amsmath}
\usepackage{soul}
\usepackage{color}
\usepackage{xcolor}
\usepackage{booktabs}
\usepackage{hyperref}
\usepackage{multirow}
\usepackage[shortlabels]{enumitem}
\usepackage{caption}
\usepackage{subcaption}
\usepackage{newtxtext,newtxmath}
\usepackage{tabularx}
\usepackage{changepage}
\usepackage{makecell}

\begin{document}

\begin{frontmatter}



\title{FedPOIRec: Privacy Preserving Federated POI Recommendation with Social Influence}

\hypersetup{pdfauthor={Vasileios Perifanis, George Drosatos, Giorgos Stamatelatos and Pavlos S. Efraimidis}}

\author[inst1,inst2]{Vasileios Perifanis\corref{cor1}}
\ead{vperifan@athenarc.gr}

\author[inst1]{George Drosatos}
\ead{gdrosato@athenarc.gr}

\author[inst1,inst2]{Giorgos Stamatelatos}
\ead{gstamatelat@athenarc.gr}

\author[inst1,inst2]{Pavlos S. Efraimidis}
\ead{pefraimi@athenarc.gr}

\affiliation[inst1]{organization={Institute for Language and Speech Processing, Athena Research Center},
            addressline={Kimmeria}, 
            city={Xanthi},
            postcode={67100}, 
            country={Greece}}
\cortext[cor1]{Corresponding author}

\affiliation[inst2]{organization={Department of Electrical and Computer Engineering, Democritus University of Thrace},
            addressline={Kimmeria}, 
            city={Xanthi},
            postcode={67100}, 
            country={Greece}}

\begin{abstract}
With the growing number of Location-Based Social Networks, privacy preserving location prediction has become a primary task for helping users discover new points-of-interest (POIs). Traditional systems consider a centralized approach that requires the transmission and collection of users' private data. In this work, we present FedPOIRec, a privacy preserving federated learning approach enhanced with features from users' social circles for top-$N$ POI recommendations. First, the FedPOIRec framework is built on the principle that local data never leave the owner's device, while the local updates are blindly aggregated by a parameter server. Second, the local recommenders get personalized by allowing users to exchange their learned parameters, enabling knowledge transfer among friends. To this end, we propose a privacy preserving protocol for integrating the preferences of a user's friends after the federated computation, by exploiting the properties of the CKKS fully homomorphic encryption scheme. To evaluate FedPOIRec, we apply our approach into five real-world datasets using two recommendation models. Extensive experiments demonstrate that FedPOIRec achieves comparable recommendation quality to centralized approaches, while the social integration protocol incurs low computation and communication overhead on the user side.
\end{abstract}



\begin{keyword}
Federated Learning \sep Privacy \sep POI Recommendation \sep Fully Homomorphic Encryption \sep Social Network
\end{keyword}

\end{frontmatter}


\section{Introduction}
Location-Based Social Networks (LBSNs), such as Foursquare, have become an everyday tool for building online social circles and sharing common interests. In particular, LBSNs provide explicit users' information, such as visits to venues \cite{cho_friendship_2011}. One of the main tasks in LBSNs services is to capture users preferences for providing useful POI recommendations \cite{cheng_fused_2012}.

On the one hand, traditional recommenders collect users check-in lists in an effort to benefit from multiple interactions and train high-quality recommendation models. On the other hand, centralized approaches tend to collect not only a user's preferences, but also individual-specific parameters such as the timestamp of a check-in and social relationships. Besides data dissemination and collection, a model's release may leak sensitive attributes. Common recommendation services mainly rely on latent factor models, i.e., matrix factorization techniques \cite{koren_matrix_2009}. Although these approaches can provide high quality recommendations, its output may lead to information leakage in both centralized \cite{narayanan_robust_2008} and collaborative learning \cite{melis_exploiting_2019}.

Privacy concerns related to users' data can be minimized by leveraging a federated learning approach \cite{mcmahan_communication-efficient_2017}. In this setting, stochastic gradient descent (SGD) iterations for a model's update are performed locally, in the data owner's device. The updates from each participating client\footnote{We use the terms user and client interchangeably.} are distributed to a parameter server, who performs an aggregation step to form the new global parameters and the process continues until model convergence.

Although federated learning is a privacy-by-design solution, it does not guarantee the prevention of information leakage from the model itself \cite{melis_exploiting_2019}. A passive parameter server, who follows the protocol but attempts to infer additional information, can deduce a user's interactions by simply observing the differences in item updates \cite{melis_exploiting_2019}, in the case of latent factors models. To avoid such a leakage, common privacy preserving techniques include the utilization of secure multiparty computation (SMC) \cite{bonawitz_practical_2017} or homomorphic encryption (HE) \cite{phong_2018_privacy} approaches for enabling a blind aggregation. In the former setting, users participated in the training process, mask their parameters with random values and enable the aggregator to cancel their masks when added together. In the latter, the aggregator can only operate on encrypted data and thus, it cannot deduce any information for the learned parameters. However, HE comes with heavy communication and computation overhead on the users' side. Hence, in this work, we utilize a SMC approach that allows the parameter server to only learn the aggregated result. 

Combining federated learning with a secure aggregation scheme can lead to a privacy preserving recommendation framework. Although latent factor models can lead to high recommendation quality, their accuracy can be further influenced by social features \cite{cheng_fused_2012}. For instance, the preferences of a user's friends can lead to better recommendations \cite{wang_location_2013}. That is, homophily suggests that individuals with common interests tend to share a friendship relationship \cite{mcpherson_birds_2001}. In federated learning, friendships are not directly available, as local data is never transmitted to an external entity.

A straightforward way to enhance personalization, and as a result the recommendation quality, is knowledge transfer among users in terms of local vectors learned by the latent factor model. For instance, matrix factorization captures the preference between a user's vector and the corresponding interacted item vectors, in an effort to provide an estimation for unobserved actions \cite{koren_matrix_2009}. In this work, we exploit the similarity of the learned vectors in a social circle after the federated computation to benefit from direct neighbors. Specifically, the transmission of a user's vector within the first neighborhood is intended to create a weighted mean vector that captures the preferences from friends for providing higher quality recommendations. However, transmitting these parameters in plain format can lead to the reconstruction of a particular user's preferences. A passive user could collect the underlying user vectors from their social circle to determine the preferences of an individual to specific items (POIs). To overcome this challenge, we propose a privacy preserving approach based on a fully homomorphic encryption (FHE) scheme, that allows the computation of the weighted mean vector by retaining data confidentiality. Our approach is generic and can also be applied to different tasks when computations with privacy concerns minimization are required.

The main contributions of this paper are summarized as follows:
\begin{itemize}
    \item We present FedPOIRec, a privacy preserving federated learning framework that can be integrated with any collaborative filtering algorithm for providing top-$N$ POI recommendations.
    \item We propose an efficient privacy preserving approach for data aggregation based on FHE computations. We benefit from social relationships and aggregate individual-level parameters over encrypted data to enhance the personalization and recommendation quality, while minimizing data exposure.
    \item We adapt two recommendation models in the federated setting: a conventional matrix factorization model optimized with Bayesian Personalized Ranking (BPR) \cite{rendle_bpr:_2009} and a sequential recommender based on a convolutional neural network (CASER) \cite{tang_personalized_2018}. To our knowledge, this is the first attempt for recommendation based on sequential patterns in federated learning. 
    \item We conduct extensive experiments on five real-world datasets. Our results show that the integration of social features enhances the recommendation quality, while FedPOIRec maintains high efficiency and minimizes privacy concerns. 
\end{itemize}

The remainder of this paper is organized as follows. Section \ref{background} describes the preliminaries of federated learning, collaborative filtering and fully homomorphic encryption. Section \ref{related_work} discusses related work in federated POI recommendation. Section \ref{FedPOIRec} presents a generic privacy preserving federated collaborative filtering algorithm which can be adapted to any top-$N$ recommender. Section \ref{ppSocial} discusses the method for integrating social features and details the operations on the FHE scheme for aggregating individual-level learned parameters in a privacy preserving manner. The experimental results are shown in Section \ref{experiments}. Finally, Section \ref{conclusion} presents the conclusions of this study.

\section{Preliminaries}
\label{background}
In this section, we first present the preliminaries of federated learning and collaborative filtering methods. Then, the concept of FHE, which we consider in our privacy preserving social integration protocol, is introduced.

\subsection{Federated Learning}
Training machine learning models is a task that traditionally requires users to transfer their data to a (trusted) third party. However, information processing and dissemination are minimized by regulations, such as the GDPR \cite{ec_gdpr_2016} and HIPAA \cite{gov_hipaa_1996}. Recently, a distributed training paradigm, federated learning \cite{mcmahan_communication-efficient_2017}, emerged which enables users to collaborate within a training computation without local data transmission. In this setting, each user locally trains a model shared from a parameter server, by utilizing the local data. Then, unlike traditional machine learning, users only share their calculated parameters, enabling privacy-by-design data utilization. The parameter server summarizes the updates received by participating clients and distributes the aggregated model to a number of randomly selected users in the next round.

Although federated learning facilitates machine learning without requiring local observations transmission, several challenges arise. Among limitations, data heterogeneity can lead to inconsistent updates, which in turn may bias the global model towards certain clients or lead to decreased convergence \cite{sattler_robust_2020}. In our learning task, users holding many observations may bias the model towards the POIs they visit. As a result, the generated model may fail to be representative of the global distribution and, consequently, it may provide low-quality recommendations to users with limited local observations. To minimize the impact of data heterogeneity, we employ a personalization and a social features fusion step, after the learning process.

Another major challenge is how the aggregation step should be performed to provide sufficient privacy guarantees. In naive federated learning, the parameter server collects several clients' updates in plain format. Therefore, private information, such as the local dataset may be deduced \cite{zhang_survey_2021}. In the case of latent factor models which are considered in this work, the parameter server can directly deduce the interacted items by observing the differences between a client's update and the previous global model \cite{melis_exploiting_2019}. To achieve privacy, cryptographic techniques, such as SMC and HE, are commonly used in the aggregation phase. In this work, we choose to integrate a SMC approach because HE introduces heavy computation and communication costs, especially when integrated with latent factor models \cite{chai_secure_2020}.

\subsection{Collaborative Filtering}
The main goal of a recommendation service is to predict a list of top-$N$ unobserved interactions based on past user's feedback \cite{sarwar_item_2001}. Users' responses fall into two main categories: rating feedback (e.g., 1-10 stars) and unary feedback, i.e., the observed instances only capture interactions without explicitly defining the user's preference \cite{desrosiers_comprehensive_2011}. Although explicit feedback formulation have been widely used for predicting unobserved ratings \cite{koren_matrix_2009}, privacy aware users may refrain from providing their response. Hence, unary feedback is easier to collect and without interrupting users' actions. In this work, we consider implicit feedback, i.e., local devices only collect user visits, without asking for a user's response.

Two common methods for the generation of a recommender are by leveraging memory-based or model-based algorithms \cite{desrosiers_comprehensive_2011}. In memory-based algorithms, the prediction is made by considering users (neighbors) whose local observations are correlated to those of the target user. Model-based approaches, on the other hand, capture latent representations of user-item interactions using a learning procedure and predict new content for the target user. However, the former approaches suffer from selection uncertainty, while model-based methods, in many cases, fail to correctly learn from latent representations due to high sparsity or limited users interactions \cite{adomavicius_toward_2005}. In this work, we first employ a model-based approach and after the training process, we benefit from direct neighbors' preferences to enhance personalization and the recommendation performance. More precisely, we make use of the latent vector that independently calculated in each user's side. Then, these vectors are used to calculate similarities among friends. Finally, the similarities are used as weights to fuse the learned parameters, thus transforming the model into a social recommender \cite{yang_survey_2014}.

\begin{figure*}[htb!]
     \centering
     \begin{subfigure}[b]{0.23\textwidth}
         \centering
         \includegraphics[width=\textwidth]{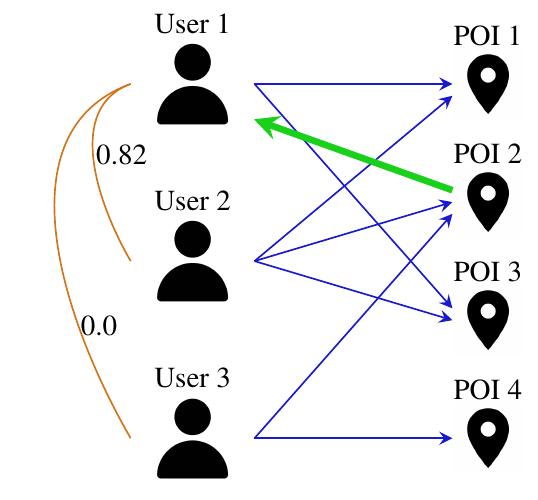}
         \caption{Memory-based.}
         \label{fig:memory_based}
     \end{subfigure}
     \hfill
     \begin{subfigure}[b]{0.32\textwidth}
         \centering
         \includegraphics[width=\textwidth]{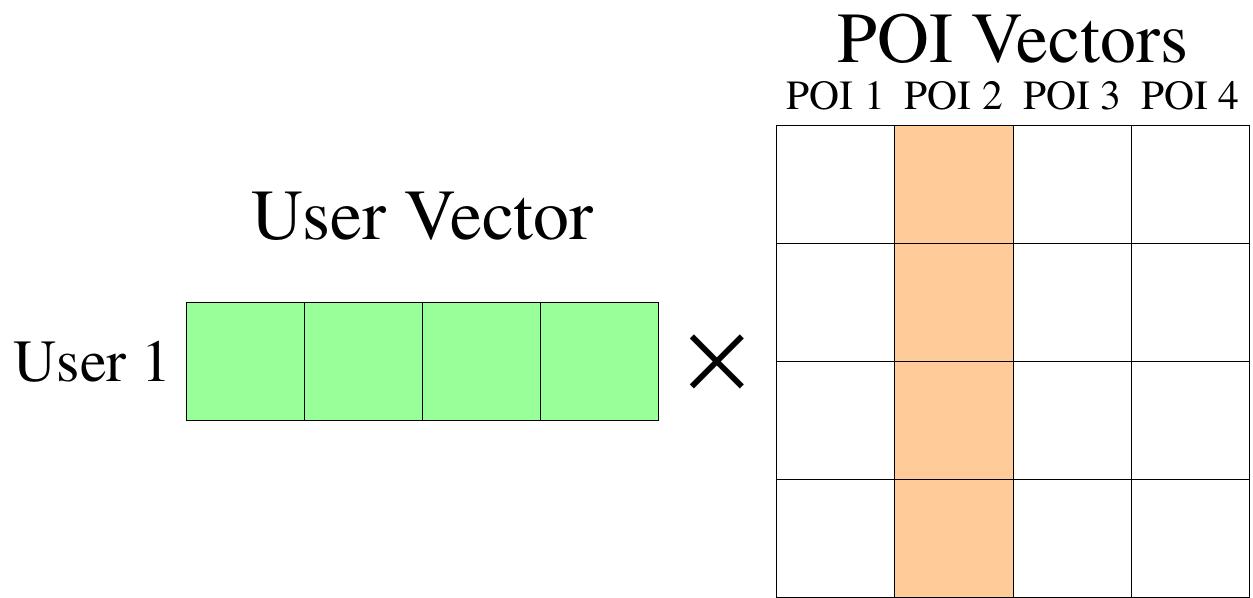}
         \caption{Model-based.}
         \label{fig:model_based}
     \end{subfigure}
     \begin{subfigure}[b]{0.43\textwidth}
         \centering
         \includegraphics[width=\textwidth]{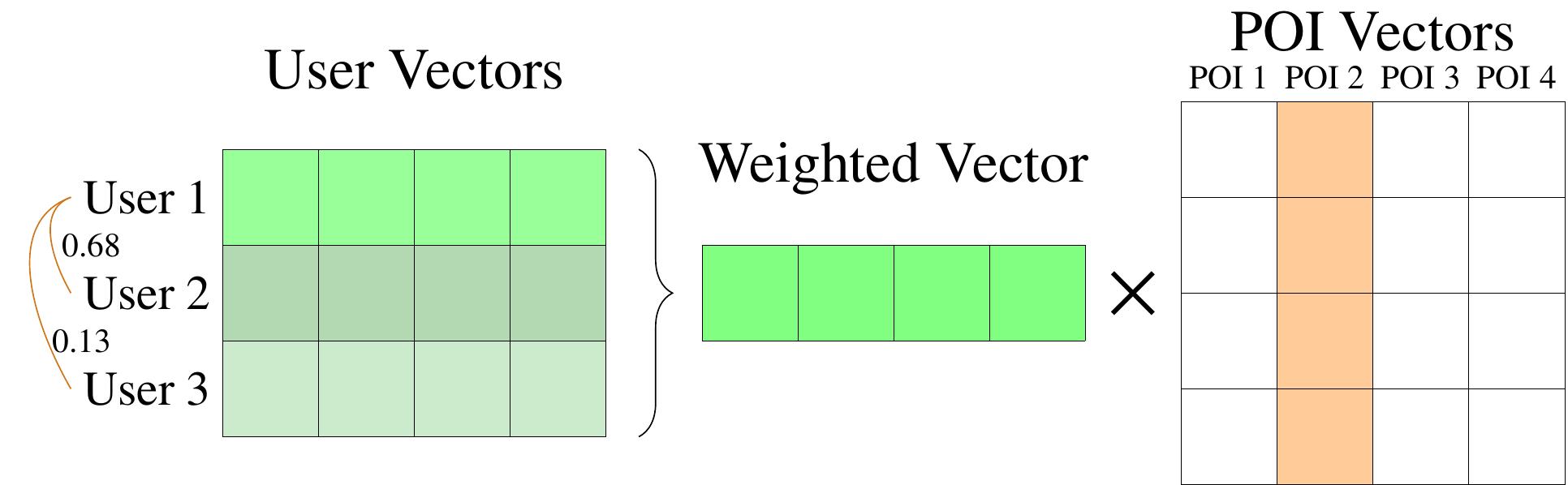}
         \caption{Our approach.}
         \label{fig:ours}
     \end{subfigure}
     \hfill
     \caption{Collaborative filtering settings.}
     \label{fig:memory_model_ours}
\end{figure*}

Figure \ref{fig:memory_model_ours} shows the recommendation generation process in a memory-based method, a simple matrix factorization (model-based) and our approach. Figure \ref{fig:memory_based} presents an illustration of the recommended POI to a user in a memory-based approach, taking into account the cosine similarities, which are equivalent to the Otsuka-Ochiai coefficient in the case of binary vectors, between the target user visits and the corresponding visits by the rest of the users. In Figure \ref{fig:model_based}, the score for each unobserved visit is calculated by the dot product between the learned user vector and the corresponding POI vector. Our approach combines the two methods in the inference stage (Figure \ref{fig:ours}). First, similarities between learned user vectors among friends are calculated and then, the target user vector is transformed considering a weighted mean of the vectors. Finally, the generated social vector is used to generate the recommendations for the target user.

\subsection{Fully Homomorphic Encryption}
\label{FHE}
In both centralized and federated model-based collaborative filtering, the learned feature vectors directly expose a user's preferences. Consequently, the social features integration, i.e., calculating the similarities between users and the corresponding aggregation, should be performed in a privacy preserving manner. In this paper, we make use of a FHE scheme, which does not require a synchronous setting. The computation can be performed by a third party as soon as encrypted data is available, without requiring the online presence of users. 

A HE scheme allows the evaluation of a function over encrypted data. Traditional HE systems, such as Paillier \cite{paillier_public-key_1999}, are \textit{partially} homomorphic, supporting either addition or multiplication on a ciphertext. Gentry \cite{gentry_fully_2009} proposed the construction of a FHE scheme, allowing the evaluation of arbitrary functions on encrypted data. Although the advances in the construction of FHE schemes, they still remain inefficient \cite{naehrig_can_2011}. For this reason, practical applications rely on their \textit{somewhat} or \textit{leveled} HE variants, which allow the evaluation of a limited or predefined number of functions for a ciphertext, respectively \cite{brakerski_2014_leveled}.

Since the parameters that need to be aggregated are floating point numbers, we utilize the Cheon-Kim-Kim-Song (CKKS) scheme \cite{takagi_ckks_2017} to generate the desired weighted mean based on a user's friends. The CKKS scheme is based on the Ring Learning with Errors (RLWE) problem \cite{lyubashevsky_ideal_2013} and allows the evaluation of a function on real numbers. Due to the approximate arithmetic nature of RLWE, noise is added to the least significant bits during encryption and hence, the decryption of a ciphertext corresponds to the original plaintext packed with extra noise. 

In this work, we exploit the homomorphic properties of CKKS to calculate cosine similarities between users over encrypted data. Then, the encrypted similarities are employed as weights for calculating the weighted mean vector. The resulting ciphertext can be only decrypted in the querier's side and therefore, the privacy of users participated in the computation is maintained. Specific details for our privacy preserving protocol are given in Section \ref{ppSocial}.

\section{Related Work}
\label{related_work}
In this section, we summarize the existing research in federated POI recommendation.

Directly related to our work are the federated systems presented in \cite{ferrara_fpl_2021, liu_prefer_2021}. Ferarra et al. \cite{ferrara_fpl_2021} extended the BPR learning algorithm \cite{rendle_bpr:_2009} to the federated setting, which we also consider in this paper. The authors proposed a plain update transmission to the parameter server by allowing users to share the local calculated vectors with a certain probability, while the learning procedure is performed considering sequential learning or selecting all clients in a training round. The authors in \cite{liu_prefer_2021} followed a standard federated setting \cite{mcmahan_communication-efficient_2017}, except that multiple parameter servers are introduced to cover geographic regions, which also collect and aggregate clients' updates in plain format. Nevertheless, both works do not study social features in the context of recommendations and the aggregation step is not performed using a secure strategy.

A first attempt that formulates on device training with neighborhood influence was presented in \cite{chen_privacy_2018}. In that distributed system, each user locally calculates the gradients of the visited venues and then, each computed parameter is propagated in plain format to remote neighbors using a random walk approach. Consequently, the remote users update their corresponding local venue vectors based on the received gradients. Although the system is fully decentralized, local gradients sharing directly exposes the preferences of a user.

In \cite{chen_practical_2020}, the collaborative filtering model is divided into a local and a global part. The local model for each user is updated by considering the computed local parameters from the k-nearest neighbors, who are found based on the geographical distance. The global part is owned by the parameter server, who collects perturbed user-venue interactions. The vector for each POI in the global profile is updated only from users who have visited this specific venue using a secure aggregation strategy \cite{bonawitz_practical_2017}. Finally, the prediction is made by considering the local and global learned parameters. The main bottleneck of the system is the communication overhead that is introduced, as users need to i) identify their k-nearest neighbors, ii) communicate with them and exchange parameters for updating the local models and iii) communicate with the parameter server for updating the global part. Moreover, the collection of user-venue interactions, although perturbed, may lead to preference leakage.

Unlike previous work on federated POI recommendation, our study presents a generic privacy preserving learning procedure, which can be integrated with any model-based collaborative filtering algorithm. In FedPOIRec, users do not outsource any of their local observations (i.e., visits to POIs) and a secure aggregation strategy is integrated to minimize the local updates exposure to the parameter server. Besides, we consider a standard federated learning scenario, without requiring all users to be available at any given time. Finally, we explore the social influence in the inference stage after the federated learning procedure. In a real-world scenario, the federated process occurs when only a fraction of users is available. Hence, the formulation of parameters transmission with privacy guarantees in the training stage incurs heavy communication and computation overhead. To avoid huge costs on the client side, we fuse learned parameters between friends after the global model's generation, using a FHE construction. Our privacy preserving protocol allows the computation of the weighted mean vector without local parameters exposure and can also be adapted to tasks besides recommendation systems.

\section{FedPOIRec System}
\label{FedPOIRec}
In this section, the federated collaborative filtering process and the privacy preserving model's aggregation technique are described without considering the social features integration part. 

\subsection{Overview}
We consider a setting where $M$ parties, each locally holds private observations who wish to jointly train a model-based collaborative algorithm. At the end of the training process, the generated model is transmitted to every participant, who performs additional training steps to achieve personalization.

The parties involved in the training process are interested in preserving the privacy of their local data and the resulting local model. To this end, we utilize a federated learning procedure enhanced with a SMC approach to keep local data and models confidential. More specifically, our learning system consists of two main entities:
\begin{itemize}
    \item \textbf{Data owners:} We consider a set $\mathcal{U}$ consisting of $M$ data owners. Each user owns a device with sufficient storage, computing capabilities and access to a network. Each client $u \in \{1, ..., M\}$ holds a local dataset $D_u$ consisting of a set of tuples in the form of $(u, i, r_{ui})$, where $i \in \{1, ..., |\mathcal{I}|\}$ is a venue's id from the global POI profile $\mathcal{I}$ and $r_{ui}$ denotes the preference of user $u$ to the venue $i$. Since, we formulate unary-response preferences, an observed interaction is defined as $r_{ui} = 1$. In the training phase, each local dataset is expanded with negative feedback by sampling unobserved interactions, thus formulating a number of non-visited POIs as $r_{ui} = 0$. In addition, the size for each local dataset $|D_u|$ can vary, while the local observations are possibly drawn from different distributions.
    \item \textbf{Parameter Server:} A parameter server is responsible for random users' selection in each global round as well as for summarizing the received updates.
\end{itemize}

Different from traditional machine learning, where an entity collects observations from multiple users, i.e., $D = \bigcup_{u \in M} D_u$, and employs a collaborative filtering algorithm, FedPOIRec utilizes a federated model-based approach. The main objective is to enable a privacy preserving training method without local data transmission.
More specifically, during local training on user $u$, no other party should learn the preference of $u$ to a specific POI $i$.

In the following subsections, the main steps of the federated training for a recommendation model generation are described. Table \ref{tab:notations} summarizes the notations and the description of parameters in the learning process of FedPOIRec.

\begin{table*}[!htb]
  \centering
  \begin{tabularx}{\textwidth}{ lX }
    \textbf{Notation}     &   \textbf{Description}\\
    \midrule
    $\mathcal{U}$   & Set of $M$ users.\\
    $\mathcal{I}$   & Set of $|\mathcal{I}|$ items.\\ 
    $D_u$   &   The local dataset of a user $u \in \{1, 2, ..., M\}$.\\
    $D_u^- = \{i \in \mathcal{I}: i \notin D_u\}$   &  Unobserved interactions for user $u$, $i \in \{1, 2, ..., |\mathcal{I}|\}$.\\
    $U_u$   & The learned vector for user $u$.\\
    $I_i$   & The learned vector for venue $i$.\\
    $U_t \subseteq \mathcal{U}, t \in T$   & Set of available participants in time step $t$.\\
    $\tilde{U}_t \subseteq U_t$ & Set of selected participants in step $t$.\\
    $W_t$   & Global model's parameters in step $t$.\\
    $[W_{t+1}^{u}]$    & Transformed model with secure aggregation.\\
    \bottomrule
  \end{tabularx}
  \caption{Notations and their description in FedPOIRec.}
  \label{tab:notations}
\end{table*} 

\subsection{Local Training}
Federated learning is defined over a number of global rounds $T$ and local epochs $E$ by an iterative process over some global parameters $W$. In each training round $t \in T$, the parameter server distributes the current global variables $W_t$ to a uniformly random subset $\tilde{U}_t \subseteq U_t \subseteq \mathcal{U}$ of available participants. Each selected participant downloads the global parameters and performs $E$ gradient descent steps using the local dataset $D_u$. Using this iterative process, the training phase is transferred on the client side and thus, users never outsource their local data.

The main challenge in federated collaborative filtering is how users can jointly build a recommendation model, which is not biased towards some clients or venues. In general, in matrix factorization, the preference of a user $u$ to an unobserved item $i$ is predicted using the dot product between the corresponding user and item vectors, i.e., $\hat{r}_{ui} = U_u \times I_i$ \cite{koren_matrix_2009}. The goal in the iterative learning process is to minimize a loss function $\mathcal{L}$ considering the user's training instances. More precisely, the learning objective can be solved using SGD by sampling a random batch of samples $D_r \subseteq D$:
\begin{equation}
\label{sgd}
    W_{t+1} = W_t - \eta \nabla \mathcal{L}(D_r, W_t)
\end{equation}
where $\eta$ is the learning rate and $\nabla$ is the gradient of the loss with respect to model parameters. Note that in a federated setting, the local user vector $U_u$ is only updated locally and reflects the preference of user $u$ to the POIs in the profile. Consequently, the user vectors are not outsourced to any entity in the system due to privacy concerns \cite{liu_prefer_2021}. 

In FedPOIRec, the iterative process of SGD is performed independently in each participating client $\tilde{u} \in \tilde{U}_t$ by randomly selecting instances from the local dataset $D_{\tilde{u}}$. In unary-response modeling, a common learning strategy is to further sample unobserved interactions \cite{rong_2008_one}. In our scenario, the local datasets only contain visits to POIs and hence, leveraging only positive interactions, would result in a biased model towards the most popular venues. Hence, the local datasets in each training iteration are expanded by selecting unobserved interactions uniformly at random.

After the local SGD iterations, each client should distribute the updated model parameters to the server for calculating the new global model. The updated model's variables $W_{t+1}^{\tilde{u}}$ from each selected client $\tilde{u}$ are transmitted to the parameter server. Upon collecting $\sum_{i \in \tilde{U}_t}W_{t+1}^i$ updates or when a waiting time passes, the server summarizes the updates to generate the global parameters for the next round. The communication between participating entities is synchronous and the process repeats until a desired level of model's quality is achieved.

\subsection{Model Aggregation}
In the naive approach of the federated computation, the parameter server collects the intermediate local parameters and generates the new global model by summarizing the received updates. In this case, the server observes all the intermediate local models, which may reveal sensitive attributes \cite{zhang_survey_2021}. Therefore, secure aggregation strategies should be utilized to minimize the potential information leakage from the transmitted updates.

The most widely employed SMC aggregation is the SecAgg scheme \cite{bonawitz_practical_2017}. Other approaches, such as \cite{bell_secure_2020, so_turbo_2021, jiang_pflm_2021}, are based on the SecAgg's construction for providing better efficiency or robustness. In this work, we utilize the SecAgg protocol to achieve a privacy preserving aggregation. Below, we summarize the main operations involved in the protocol. Proposing new ways for aggregation is beyond the scope of this paper.

The main idea behind SecAgg \cite{bonawitz_practical_2017} is that users agree on one-time masks for each calculated value. More precisely, each pair of users agree on random seeds and each user locally computes the masks based on the random seed using a pseudo random generator (PRG). Finally, the calculated local parameters on a user $\tilde{u}$ after applying SGD, are transformed using the random variables by:
\begin{equation}
\label{secagg_single_mask}
     [ W_{t+1}^{\tilde{u}} ] := W_{t+1}^{\tilde{u}} + \sum_{i:\tilde{u} < i} PRG(\tilde{u}, i) - \sum_{i:\tilde{u} > i}PRG(\tilde{u}, i)
\end{equation}

The server collects the blurred parameters and correctly calculates the sum of the updates by canceling the introduced masks. To ensure fault tolerance, additional masks are randomly generated in each user's side, which are then secret shared with other participants using a t-out-of-N secret sharing technique, thus further transforming the masked parameters calculated in equation \ref{secagg_single_mask} by:
\begin{equation}
\label{secagg_double_mask}
    [ W_{t+1}^{\tilde{u}} ] := [ W_{t+1}^{\tilde{u}} ] + PRG(\tilde{u})
\end{equation}

When a user drops, the server communicates with t participants to cancel the associated masks. Note that the SecAgg scheme can effectively handle up to $({|U_t|}/{2}) - 1$ user failures.

Although, SMC approaches introduce some additional communication cost between users, heavy computation tasks and storage requirements are not required compared to HE approaches. For instance, in the simplest form of a latent factor model, users should encrypt the whole item profile, i.e., a $d \cdot |\mathcal{I}|$ matrix, where $d$ is the latent dimension and $|\mathcal{I}|$ is the number of items in the profile \cite{chai_secure_2020}. On the other hand, utilizing a SMC approach, users only communicate to exchange a small parameter and perform a simple transformation on the latent matrix. Therefore, HE approaches are not suitable for recommendation systems and thus, SMC is selected as the building block of the secure aggregation strategy.

\subsection{FedPOIRec Learning Process}
Putting it all together, the privacy preserving learning process of FedPOIRec is given in Algorithm \ref{fedpoirec_alg}. The number of global and local rounds $T$ and $E$, respectively, the global parameters in the first round $W_0$, the client set $\mathcal{U}$, the fraction of online participants $C$ to consider in a global round and the number of unobserved interactions $numNeg$, which is a special parameter in the case of unary-feedback for collaborative filtering, are given as input. Then, the parameter server samples the available participating clients at time step $t$, selects uniformly at random at least three online users for local training and transmits them the current global parameters (lines \ref{fedpoirec_alg:start_sampling} - \ref{fedpoirec_alg:end_sampling}). The constraint of three available clients is introduced to prevent a passive entity from deducing information in the next round. For instance, consider a scenario with two participants, one of whom is a passive entity and selected in consecutive rounds. The difference between the aggregated model in the next round and the local model calculated in the precious step, discloses the local weights of the honest participant.

Each selected client receives the current global model and calculates the set of unobserved items $D_{\tilde{u}}^{-}$. Then, $E$ gradient descent steps are performed using the local observations expanded with unobserved items, which are selected uniformly at random (lines \ref{fedpoirec_alg:start_learning} - \ref{fedpoirec_alg:end_learning}). Finally, the updated local model is blurred using one-time masks and is transmitted to the parameter server for aggregation. In the last step of the learning procedure, the parameter server summarizes the updates and generates the new global model $W_{t+1}$ (line \ref{fedpoirec_alg:new_model}).

\begin{algorithm}
\caption{FedPOIRec}
\label{fedpoirec_alg}
    \hspace*{\algorithmicindent} \textbf{Input:} $T, E, W_t, \mathcal{U}, C,$ numNeg \\ 
    \hspace*{\algorithmicindent} \textbf{Output:} $W_{T}$
    \begin{algorithmic}[1]
    \For {$t=0$ to $T$}
        \State sample online clients $U_t \subseteq \mathcal{U}$ \label{fedpoirec_alg:start_sampling}
        \State totalParticipants = max(|$U_t$| $C$, 3), |$U_t$| $\geq$ 3 \Comment{Get the number of participating clients.}
        \State $\tilde{U}_t$ = random($U_t$, totalParticipants)
        \State transmit $W_t$ to the selected participants  \label{fedpoirec_alg:end_sampling}
        \For {each $\tilde{u}$ in $\tilde{U}_t$}
            \State $D_{\tilde{u}}^{-}$ = $\{i \in \mathcal{I}: (u, i, 0) \not\in D_{\tilde{u}}\}$  \label{fedpoirec_alg:start_learning} \Comment{Generate dataset containing unobserved interactions.}
            \For {$e=1$ to $E$}
                \State unobserved = random($D_{\tilde{u}}^{-}$, numNeg) \Comment{Sample negative interactions.}
                \State $D_{\tilde{u}} = D_{\tilde{u}}$ $||$ unobserved \Comment{Expand local dataset with negative feedback.}
                \State Get $W_{t+1}^{\tilde{u}}$ using SGD (equation \ref{sgd}) \label{fedpoirec_alg:end_learning}
            \EndFor
            \State $[ W_{t+1}^{\tilde{u}} ]$ := SecAgg  (equation \ref{secagg_double_mask}) \Comment{Transform parameters using secure aggregation.}
            \State distribute $[ W_{t+1}^{\tilde{u}} ]$ to server
        \EndFor
        \State $W_{t+1} = \sum_{i=0}^{|\tilde{u}|} [ W_{t+1}^{i} ] \: / \: |u|$ \Comment{Secure aggregation.} \label{fedpoirec_alg:new_model}
    \EndFor
    \State \Return $W_{T} = W_{t+1}$
    \end{algorithmic}
\end{algorithm}

After convergence, the training phase ends and the global model is finalized. At this stage, the federated collaborative filtering model may provide sufficient recommendations to each user. Due to the nature of federated learning, inevitably, some users were not selected in recent global updates. For instance, some clients may were offline or the random process did not select them. Hence, the learned parameters may fail to provide high quality recommendations to those users. To overcome this limitation, the participating clients run a number of local epochs to further minimize the local objective.

\subsection{Privacy Analysis}
In our learning scenario, we assume that participating entities, i.e., the data owners and the parameter server, are honest in terms of correctly performing SGD iterations and the corresponding aggregation, respectively, but they try to infer additional information for a participating user $u$. In addition, we assume that the parameter server and the participating clients do not collude.

The adopted secure aggregation strategy prevents the server from directly deducing information regarding a single user's update. Although in each training round the sum of the plain updates is correctly calculated, the intermediate results do not leak a specific user's interactions. In a similar manner, the participating clients in the next round cannot deduce information regarding the clients who updated the global model in the previous round. A special case, where data leakage can be caused by a passive user is when the parameter server selects only two clients in two consecutive rounds, as previously discussed. This limitation is almost unlikely to happen in federated settings given that there is a huge number of participants \cite{bonawitz_towards_2019} and is further alleviated by adding the constraint of selecting at least three clients (Algorithm \ref{fedpoirec_alg}). It is worth noting that an attacker's advantage for deducing information from an aggregated global model is almost negligible and decreasing by the number of clients who participate in a global round \cite{nasr_comprehensive_2019}.

To summarize, the utilization of a federated setting enhanced with a secure aggregation strategy based on SMC, minimizes privacy concerns regarding the identification of a user's preferences as it enables an aggregation of seemingly random parameters, without requiring users to transmit any of their local data.

\section{Privacy Preserving Social Features Integration}
\label{ppSocial}
In this section, we describe our protocol for enhancing the recommendation quality by computing a weighted average of user vectors in a social circle. First, we present an overview of our approach on plaintext data and then, we proceed to describe our privacy preserving protocol leveraging the CKKS FHE scheme.

\subsection{Overview}
Intuitively, a model-based collaborative filtering algorithm integrates the correlation between users (and as a result, friends) in the training stage. In the centralized setting, the learning algorithm has access to model parameters, including user vectors and adapts to clients with similar behavior. In contrast, in the considered scenario, the local user vectors are never transmitted from clients and consequently, the learned model may fail to correctly capture the similarities between users. Hence, we argue that transferring knowledge between learned parameters in a social circle in terms of user vectors, after the federated learning process, leads to recommendation quality enhancement. To this end, the local vectors are fused using a weighted averaging aggregation. More specifically, similarities among friends are calculated using the users' latent representations, which reflect the preferences to POIs. The similarity between two users is calculated from the cosine measure, which is widely employed in memory-based collaborative filtering methods \cite{desrosiers_comprehensive_2011}. Formally, the cosine similarity between two vectors $a$ and $b$ is defined as:
\begin{equation}
    \label{plain_cosine}
    \cos(a, b) = \dfrac{a \times b}{\|a\| \: \|b\|}= \dfrac{\sum_{i=1}^{d} a_i \: b_i}{\sqrt{\sum_{i=1}^{d} a_i^2} \: \sqrt{\sum_{i=1}^{d} b_i^2}}
\end{equation}
with $d$ the vector's dimension and $\|a\|$, $\|b\|$, the euclidean norm of vector $a$ and $b$, respectively.

The similarity between users is used as weight to generate a weighted mean vector, which replaces the local user latent representation, i.e., social features are integrated into the local model. More specifically, the local learned user vector is transformed by:
\begin{equation}
\label{plain_weighted}
    \text{WeightedVector}_\text{u} = \dfrac{\sum_{v \in \mathcal{F}_u} w\left(u,v\right) \: U_v}{\sum_{v \in \mathcal{F}_u} w\left(u,v\right)},
\end{equation}
where $\mathcal{F}_u$ is the set of friends for a user $u$ with $u \in \mathcal{F}_u$ and $w\left(u, v\right)$ denotes the cosine similarity between users $u$ and $v$ with respect to the local user vectors. In our scenario, there are two master entities, the parameter server who coordinated the federated learning procedure and is not aware of social connections and the social network service which only knows the friendships set $\mathcal{F}_u$ for each user. Note that the local parameters learned in each client are not leaked to any other party. Our goal, is to enable a privacy preserving preference fusion by calculating similarities between friends without revealing any information regarding the local parameters.

We choose to only integrate the friends' vectors as a way for approximating a user's nearest neighbors in the whole network. Given that participating users in federated learning may be millions \cite{bonawitz_towards_2019}, computing similarities among all users introduce heavy communication and computation overhead.

The integration of social features may be performed directly in the training phase. For instance, SBPR \cite{zhao_2014_SBPR}, which adapts social information on top of BPR \cite{rendle_bpr:_2009}, leverages friendships by further sampling unobserved items that are interacted by the friends of the target user. However, the approach of modeling social connections in the training stage of federated learning is not straightforward. First, exchanging local observation violates the principle of data locality. Second, the availability of devices varies and thus, real-time parameters distribution even if local observations are secured under a privacy preserving mechanism, cannot be ensured. Besides, the communication and computation overhead may be restrictive and therefore, we leave the adaptation of social features integration in the training stage for future work.

\subsection{Problem Definition}
After the federated computation, each user owns a local user vector $U_u$. We study the influence of social relationships on recommendations by generating a weighted mean user vector, which combines the personal taste of a user with the preferences of their friends, regarding the POIs in the global item profile. By incorporating social features, we aim at providing higher quality recommendations. 

The goal for each user $u$ is to compute a weighted average by incorporating the local user vector and the vectors of the direct neighbors. The main objective is to provide privacy guarantees to both queriers and friends, i.e., the local user vectors should not be disclosed to any party except for the corresponding owner. The computed local vector directly reflects the preferences of a user and thus, it should not revealed.

To overcome this challenge and maintain both high efficiency and privacy, we propose a privacy-preserving computation based on the CKKS scheme. From the definition of cosine similarity (equation \ref{plain_cosine}), the number of multiplications and additions required for the calculation is fixed. Hence, we use the leveled variant of CKKS for better efficiency. Our protocol is practical since direct connections and synchronization between users are not required, while the operations for computing the desired weighted mean vector are performed over encrypted data.

\subsection{System Model}
The solution is based on a single server, called cryptoEvaluator, that operates on encrypted data. Note that cryptoEvaluator may be equivalent to the parameter server from the federated computation. Below, we summarize the entities involved in a weighted mean vector computation:
\begin{itemize}
    \item \textbf{Querier:} A user who participated in the federated learning process and owns a local user vector. The querier initiates a weighted mean vector computation by outsourcing the local user vector in an encrypted form to cryptoEvaluator.
    \item \textbf{Friends:} Users who own local vectors (participants in the federated computation) and are direct neighbors of a querier. After receiving a weighted mean computation plan, they outsource their local vectors to cryptoEvaluator, encrypted under the querier's public key. 
    \item \textbf{cryptoEvaluator:} A third party who collects encrypted data and performs evaluation operations on ciphertexts.
\end{itemize}

On a higher level of the protocol, each user (querier) initiates a weighted mean vector computation plan. The local vector and the vectors from friends are outsourced in an encrypted form using the querier's public key to cryptoEvaluator. After computing the similarities over encrypted data, the resulting ciphertexts are utilized from cryptoEvaluator as weights for calculating a weighted mean of the outsourced vectors. Note that the operations are always performed on encrypted data and therefore, cryptoEvaluator cannot deduce any information. Below, we present the complete protocol, which is secure against (non-colluding) passive parties.

\subsection{The Privacy Preserving Social Integration Protocol}
In this section, we describe our privacy preserving mean vector computation protocol from a user's friends. We emphasize that at the end of the protocol the input vectors should not be leaked to any party (except for the vector owner), the identities of the querier's friends should not be disclosed to any party (except for the querier) and the output should be decryptable only under the secret key of the user who initiated the weighted mean vector computation.

\begin{table*}[hbt!]
    \centering
    \begin{tabularx}{\textwidth}{lllX}
        Operation   & Input &   Output  &   Description\\
        \midrule
        ContextGen  &   $params$  &  $context$  &   Public parameters generation\\
        KeyGen  &   $context$ &  $PK, SK$ &   Public-secret key pair generation\\
        EvalKeyGen  &   $context, SK$ &   $EvalK$ &   Evaluation key generation\\
        Encrypt &   $PK, m$   &   $c$ &   Ciphertext $c$ generation from plaintext $m$\\
        Decrypt &   $SK, c$   &   $m'$    &   Plaintext $m'$ recovery from ciphertext $c$\\
        EvalAdd &   $c, x$    &   $c'$    &   Addition operation\\
        EvalMul &   $EvalK_{Mult}, c, x$ &   $c'$    &   Multiplication operation\\
        EvalSum &   $EvalK_{Sum}, c$  &   $c'$    &   Sum operation\\ 
        EvalDot &   $EvalK_{Sum}, EvalK_{Mult}, c, x$ &   $c'$    &   Dot product operation\\
        EvalRot   & $EvalK_{Rot}, c, index$   &   $c'$    &   Rotation operation\\
        \midrule
  \end{tabularx}
  \caption{Operations included in the complete privacy-preserving protocol. Ciphertext and plaintext values are packed into a vector form.}
  \label{tab:schemeAlgs}
\end{table*}

In the following subsections, we summarize the four main phases of our protocol. The scheme can be extended to the Nearest Neighbors Search problem \cite{roussopoulos_nearest_1995} as well as adapted to a network without friendships. Inevitably, HE introduces additional communication and computation overhead. Hence, social information is intended to approximate a user's k-nearest neighbors in the whole network. Table \ref{tab:schemeAlgs} summarizes the operations needed for calculating the weighted averaged vector in our protocol. Note that the opetations on FHE schemes require evaluation keys that need to be transferred to the entity which performs computations on encrypted data \cite{albrecht_homomorphic_2019}.

\subsubsection{Phase 1 -- Key Generation}
\begin{figure*}[htb!]
     \centering
     \includegraphics[width=0.7\textwidth,height=0.7\textheight,keepaspectratio]{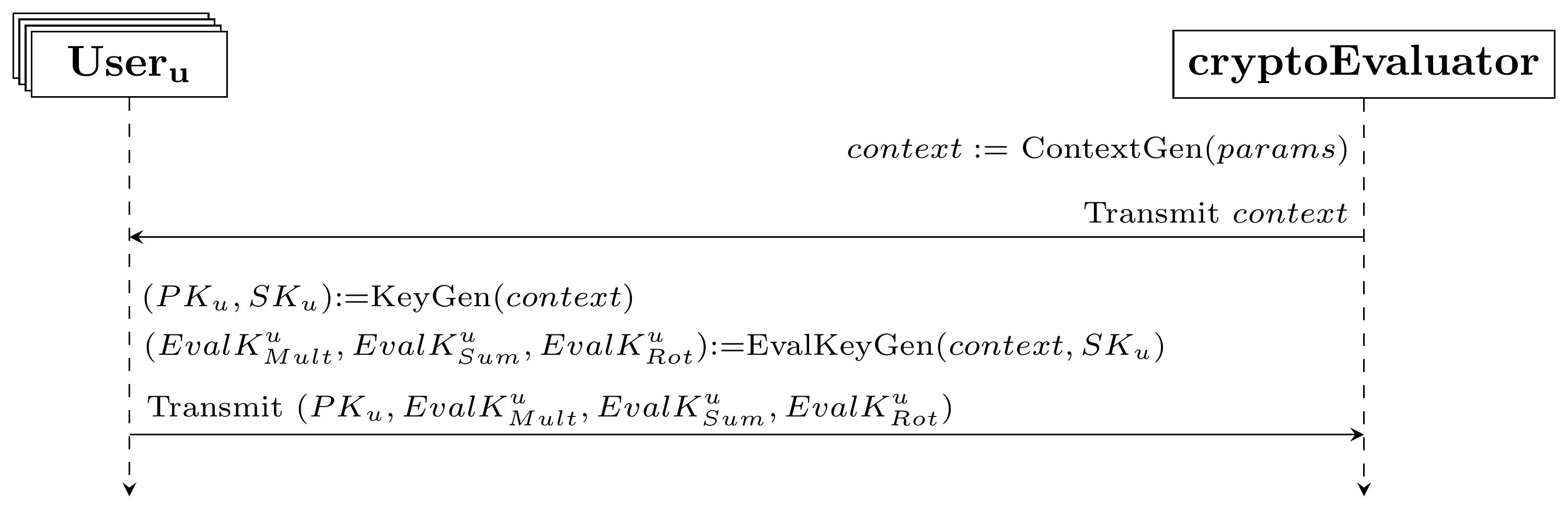}
     \caption{Phase 1 overview.}
     \label{fig:phase1_fhe}
\end{figure*}

In the key generation phase, cryptoEvaluator and clients agree on a context, i.e., public parameters. At the end of this phase, users generate key pairs as well as evaluation keys, which will be later transmitted to cryptoEvaluator for operating on encrypted data. Figure \ref{fig:phase1_fhe} summarizes the operations and the communication on the cryptoEvaluator and user side. More precisely, the key initialization phase consists of the following steps:

\begin{enumerate}
    \item CryptoEvaluator runs a setup process by ContextGen($params$) to create a context which includes the public parameters. The context is transmitted to users who participated in the federated learning process.
    \item Each user, after receiving the public parameters, generates a public-secret key pair ($PK_u, SK_u$), using the Keygen($context$) algorithm. In addition, a tuple of evaluation keys ($EvalK_{Mult}^{u}, EvalK_{Sum}^{u}, EvalK_{Rot}^{u}$) are independently generated by EvalKeyGen($context, SK_u$) for enabling cryptoEvaluator to operate on encrypted data; $EvalK_{Mult}$ is used for performing multiplications, $EvalK_{Sum}$ for calculating the sum of a vector and $EvalK_{Rot}$ for rotating on encrypted data, which is discussed in a later part of this section.
    \item Each client transmits a tuple of keys ($PK_u, EvalK_{Mult}^{u}, EvalK_{Sum}^{u},$ $EvalK_{Rot}^{u}$) to cryptoEvaluator. Note that these keys should not reveal anything for the underlying secret key $SK_u$ nor the plaintext values by observing data encrypted under $PK_u$.
\end{enumerate}

\subsubsection{Phase 2 -- Mean Vector Computation Announcement and Encrypted Data Transmission}
Users may initialize a mean vector computation plan after their key generation process. First, the querier $u$ prepares an encrypted vector $CU_u := \text{Encrypt(}PK_u,$ $U_u$). Depending on the similarity measure, additional parameters may be included in the encrypted local user vector. In this work, we employ the cosine measure (equation \ref{plain_cosine}) which calculates the similarity by the inner product between two vectors and the euclidean norm of each vector.

In a naive plaintext communication scenario for calculating the cosine similarity, three main operations should be performed: i) dot product ii) multiplication and iii) division. The latter operation requires inferring whether a ciphertext is invertible and then, in our case, finding the inverse of the euclidean norm based on the received encrypted vector.

The sub-tasks for performing division over encrypted data incur significant cost and consequently, are replaced by transferring the calculation of the inverse euclidean norm on the user's side. This way, each user calculates the inverse euclidean norm using the (plain) local vector and then encrypts the generated value, enabling cryptoEvaluator to correctly calculate cosine similarities by performing multiplication. Hence, we ask each user who participates in a weighted vector computation to locally calculate the encryption of the inverse euclidean norm as $CU_{u}^{Norm} := \text{Encrypt(}PK_u, Norm$), where $Norm = 1 \: / \: \|U_u\|$, to provide a division free operation in a later stage of the protocol.

The resulting ciphertext tuple ($CU_u, CU_u^{Norm}$), a random unique identifier $UUID_u$ and a waiting time threshold $WT_u$ are transmitted to cryptoEvaluator, who associates the parameters with the key tuple ($PK_u,EvalK_{Mult}^{u},$ $EvalK_{Sum}^{u},$ $EvalK_{Rot}^{u}$). 

The computation plan, consisting of the triplet ($PK_u, UUID_u$, $WT_u$) is announced to the friends of the querier $u$, e.g., through a social network service. The service inspects the plan and aborts it when the number of friendships is less than two because at the end of the computation, a single friendship will lead to the leakage of the friend's vector to the querier. The waiting time parameter determines the period in which users can send encrypted data under $PK_u$ associated with a specific $UUID_u$ to cryptoEvaluator. The plan allows a querier's friends to control the dissemination of their encrypted local vectors. For instance, a user may decide not to participate due limited resources. 

When a querier's friend has been informed and accepted the computation plan, the encryption processes of the local vector and the inverse euclidean norm are initiated using the $PK_u$. The encryption operations may be launched in compliance with federated learning job schedules \cite{bonawitz_towards_2019}, i.e., when the local device is in charge mode, idle and connected to a network.

\begin{figure*}[htb!]
     \centering
     \includegraphics[width=0.8\textwidth,height=0.8\textheight,keepaspectratio]{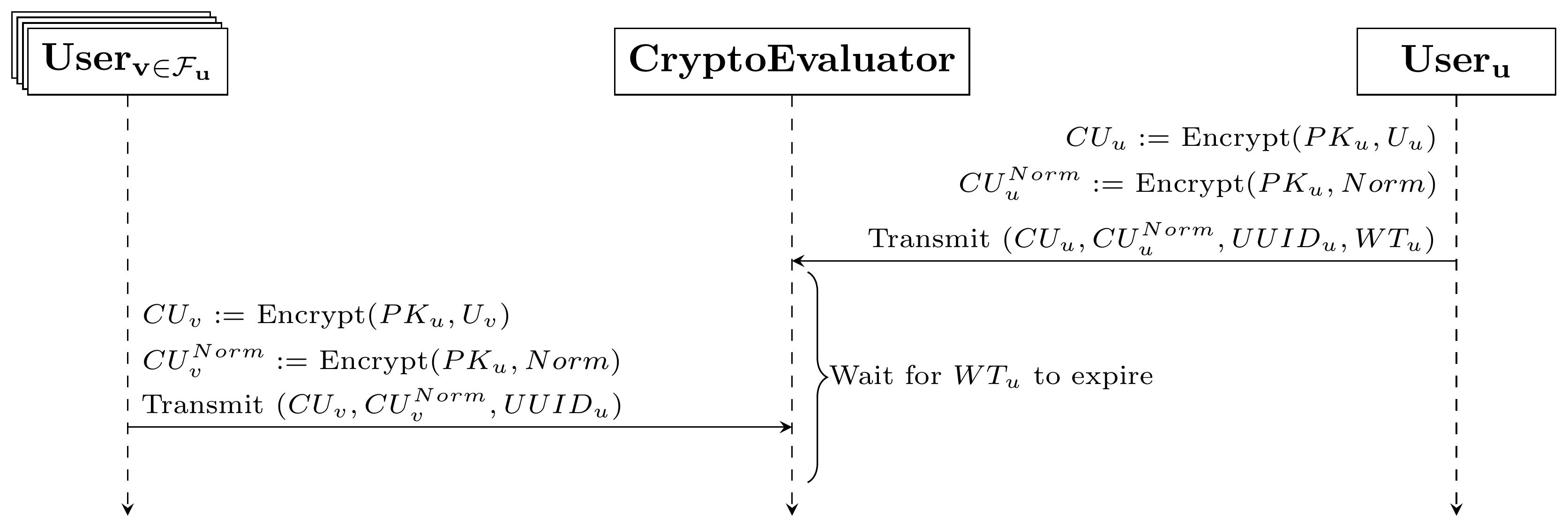}
     \caption{Phase 2 overview.}
     \label{fig:phase2_fhe}
\end{figure*}

The resulting ciphertexts along with the corresponding $UUID_u$ are transmitted to cryptoEvaluator through anonymous connections, e.g., using Tor's onion routing \cite{dingledine_tor_2004}. The anonymous communications are intended to prevent cryptoEvaluator from reconstructing the social graph from clients, e.g., by inspecting client IP addresses. The received encrypted data are appended into a $c_{vectors}^u$ and $c_{norms}^u$ matrices, respectively, which will be later used for calculating encrypted similarities and the final weighted mean vector. Figure \ref{fig:phase2_fhe} summarizes the encryption operations on the querier $u$ and their friends $v \in \mathcal{F}_u$.

\subsubsection{Phase 3 -- Similarity Computation}
Once the waiting time $WT_u$ has passed, the process continues for calculating the desired weighted average. First, similar to the previous inspection regarding the number of friends, cryptoEvaluator reviews the number of received encrypted parameters and continues the process only when more than two friends has responded to the computation plan. The first step in the computation is the calculation of cosine similarities between the querier's vector and the friends' vectors. The received encrypted vectors and the corresponding euclidean norms are sequentially processed by:
\begin{equation}
\label{evaldot}
    \begin{aligned}
        \text{Edot}\left(U_u, U_v\right) & = \text{EvalDot}\left(EvalK_{Sum}^u,EvalK_{Mult}^{u}, CU_u,CU_v\right) \\ & =
        \text{E}\left(\sum_{i=1}^{d} CU_{u_i} \: CU_{v_i}\right) =
        \text{E}\left(U_u \times U_v\right),
    \end{aligned}
\end{equation}
\begin{equation}
\label{evalmul}
    \begin{aligned}
        \text{Enorm}\left(U_u, U_v\right) & = \text{EvalMul}\left(EvalK_{Mult}^{u}, CU_{u}^{Norm}, CU_{v}^{Norm}\right) \\ & = 
                         \text{E}\left(\dfrac{1}{\sqrt{\sum_{i=1}^{d} {U_{u_i}}^2} \: \sqrt{\sum_{i=1}^{d} {U_{v_i}}^2}}\right) =
                         \text{E}\left(\dfrac{1}{\|U_u\| \: \|U_v\|}\right).
    \end{aligned}
\end{equation}
Multiplying (\ref{evaldot}) by (\ref{evalmul}), cryptoEvaluator generates the cosine similarity between the querier $u$ and the friend $v$:
\begin{equation}
\label{enc_cosine}
    \begin{aligned}
        \text{Ecos}\left(u, v\right) & = \text{Edot}\left(U_u, U_v\right) \: \: \text{Enorm}\left(U_u, U_v\right) \\ & = E\left(U_u \times U_v\right) \: \: \text{E}\left(\dfrac{1}{\|U_u\| \: \|U_v\|}\right)
               = E\left(\dfrac{U_u \times U_v}{\|U_u\| \: \|U_v\|}\right).
    \end{aligned}
\end{equation}

The computation of the cosine similarity for two vectors results in a value to the range $[-1, 1]$. Performing the weighted average using negative values may result in breaking the learned parameters for the querier and subsequently, in low quality recommendations. Hence, a normalization step should be performed. The most common approach for normalization is to transform the resulting output to the range $[0, 1]$. In our scenario, the computed similarities are only used as weights and therefore, only an addition is performed to the encrypted outputs by a scalar of $1$, thus transforming the resulting values to the range $[0, 2]$.

\subsubsection{Phase 4 -- Weighted Mean Computation}
After calculating the cosine similarities between the querier's vector and their friends' vectors, the resulting ciphertexts are used as weights for generating a weighted average. In a plaintext scenario, i) each received vector is multiplied with the calculated cosine similarity (i.e., a scalar), ii) the resulting vectors as well as the weights are added together and finally, iii) a division operation between the sum of the weighted vectors and the sum of weights is performed.

For efficient computations over ciphertexts, especially in a vector format, Single Instruction Multiple Data (SIMD) style operations are commonly used \cite{smart_fully_2014}. Another strategy is to independently encrypt each value in a vector. However, this method will result in a huge ciphertext expansion, which further leads to huge storage and communication costs. Hence, we use a SIMD-like structure to transform a vector to a single ciphertext, which enables computations to be performed in parallel. 

The first step for computing a weighted mean, is to multiply the encrypted cosine similarities with the corresponding vectors. However, SIMD operations over ciphertexts are performed element-wise. The cosine similarity is calculated on encrypted vectors (equation \ref{enc_cosine}) and therefore, the resulting output is an encrypted SIMD-like structure, which contains the desired cosine similarity in the first slot (index) of the vector, while the rest of the slots contains values close to zero. Subsequently, by multiplying the encrypted cosine with the corresponding encrypted vector, leads the resulting output with a correct value in the first slot and values close to zero in the rest slots, thus breaking correctness.

To overcome the above limitation, rotation, i.e., shifting the value from a slot to other slots in a vector \cite{halevi_faster_2018}, is introduced on the encrypted cosine similarity. More precisely, cryptoEvaluator shifts the first slot to the right to clone the encrypted similarity to every other slot in the dimension $d$ using the rotation key, $EvalK_{Rot}^u$. The rotation operation is almost as expensive as multiplication and is the main bottleneck in our approach.
\begin{algorithm}[htb!]
\caption{Weighted Vector Computation}
\label{eweighted_alg}
    \hspace*{\algorithmicindent} \textbf{Input:} vector $\langle$ Ciphertext $\rangle$ eVector, vector $\langle$ Ciphertext $\rangle$ eCos, $d$ \\
    \hspace*{\algorithmicindent} \textbf{Output:} vector $\langle$ Ciphertext $\rangle$ eResult
    \begin{algorithmic}[1]
    \State vector $\langle$ Ciphertext $\rangle$ eCosRot, vector $\langle$ Ciphertext $\rangle$ tmp
    \For{$i=1$ to $d - 1$}
        \If {$i=1$}
            \State tmp := EvalRot(eCos, -1) \Comment{Right shift} \label{eweighted_alg:start_rot}
            \State eCosRot := eCos + tmp \Comment{SIMD addition}
        \Else
            \State tmp := EvalRot(tmp, -1) 
            \State eCosRot := eCosRot + tmp \label{eweighted_alg:end_rot}
        \EndIf
    \EndFor
    \State vector $\langle$ Ciphertext $\rangle$ eResult := eVector $\ast$ eCosRot \label{eweighted_alg:mul} \Comment{SIMD multiplication}
    \State \Return cResult
    \end{algorithmic}
\end{algorithm}

The process for calculating a weighted vector on encrypted data using the resulting cosine similarity calculated in Phase 3 is described in Algorithm \ref{eweighted_alg}. To achieve the cloning operation, a rotation procedure is introduced for each latent dimension $d$. For each rotation, a SIMD addition is performed between the rotated vector and the previous rotation (lines \ref{eweighted_alg:start_rot} - \ref{eweighted_alg:end_rot}). Note that the addition operation cancels the noise introduced from the CKKS encoding. 
The resulting ciphertext contains the value of the cosine similarity in each slot. Finally, a SIMD multiplication between the calculated vector after rotations and the corresponding encrypted user vector (line \ref{eweighted_alg:mul}) is performed.

\begin{figure*}[htb!]
     \centering
     \includegraphics[width=0.7\textwidth,height=0.7\textheight,keepaspectratio]{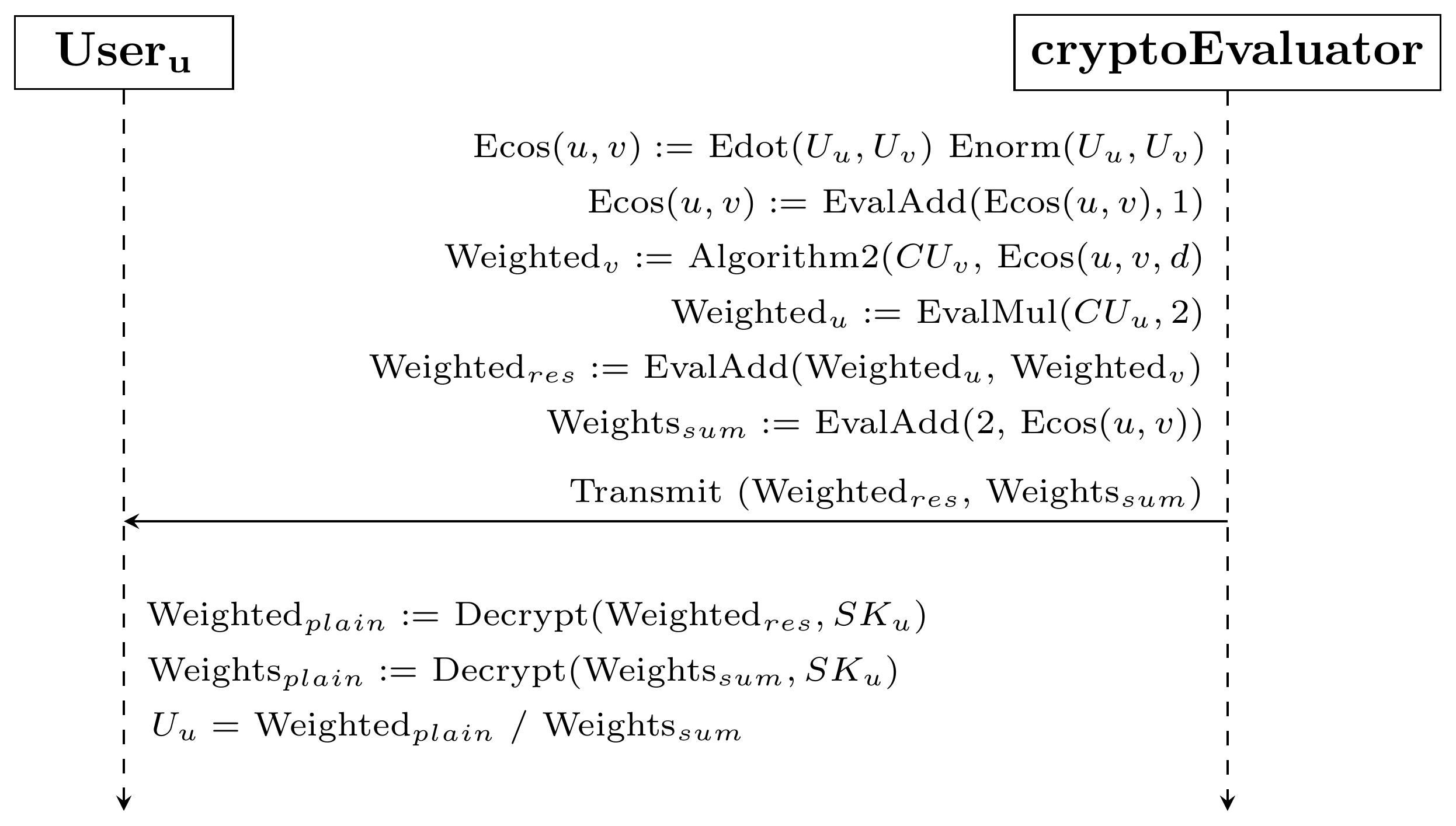}
     \caption{Phases 3-4 overview.}
     \label{fig:phase3_fhe}
\end{figure*}

After performing the rotation and multiplication operations for each received vector, the generated weighted vectors as well as the resulting cosine similarities are added together. Note that a weighted average computation requires a division (equation \ref{plain_weighted}), which is an expensive operation as already discussed. Hence, both the numerator and denominator are transferred to the querier, who performs division on plaintext values after decryption. Figure \ref{fig:phase3_fhe} shows the computations involved in the cryptoEvaluator's side as well as the decryption operation on the querier's $u$ side for generating a weighted mean user vector based on a single friendship.

\subsection{Privacy Analysis}
In our privacy-preserving mean vector computation protocol, similar to the federated setting, we assume that the participants and cryptoEvaluator are honest in terms of their computations, but they may try to infer additional information regarding the local learned parameters of a user $u$. The learned parameters leak the preferences of the target user and consequently, a local vector exposure leads to local dataset deduction.

The transmitted parameters for a computation plan are always in an encrypted form under the querier's public key $PK_u$. Hence, cryptoEvaluator cannot decrypt any intermediate result. However, a different privacy challenge arises. Although cryptoEvaluator cannot deduce any information from the outsourced ciphertexts due to the CPA-security guarantee of the CKKS scheme \cite{li_security_2021}, it can still learn the friendships graph by collecting the underlying IP addresses in a weighted mean vector computation plan. This way, it can reconstruct the social graph for each querier and by merging the collected sub-graphs, the whole social graph is leaked in plain format. Consequently, a linking attack that may identify the individuals can be performed. To alleviate this threat, first, the friends of a querier control the dissemination of their local vectors and second, the encrypted vectors distribution to cryptoEvaluator is performed through anonymous communications. This ensures the anonymity of the participating clients and therefore, cryptoEvaluator cannot execute the identified linking attack.

Except for privacy guarantees against cryptoEvaluator, privacy against passive users should be also ensured. Recall that queriers are aware of the their local vectors and the final output after the weighted mean vector computation. Although, in the general case, the querier is not able to deduce the friends' local vectors, there is a direct privacy leakage when the computation plan only involves a single friend. In this case, the querier can easily compute their friend's local vector based on the resulting output transmitted from cryptoEvaluator and consequently, to determine the friend's exact preferences. To prevent such a leakage, we have introduced inspection operations before the announcement of the computation plan and when the computation over encrypted data is about to be utilized. First, the social network service aborts the computation when a querier has less than two friends and second, cryptoEvaluator aborts it when only a single encrypted vector is received for a computation plan. This way, the computation requires the transmission of at least two friends' vectors to be utilized. The final aggregated output blurs the preferences of a single friend and therefore, the leakage of preferences is minimized.

Summarizing, the outsourced local vectors are protected under the CPA-security of the underlying FHE scheme, i.e., no party can deduce any information about the plaintext data by observing the corresponding ciphertexts, 
which are encrypted under a $PK_u$, except for the entity that holds the corresponding secret key $SK_u$. A potential linking attack is alleviated using anonymous communications and an information deducing attack is minimized by checking operations before execution. Hence, our protocol provides high privacy guarantees against non colluding passive entities.

\section{Experiments}
\label{experiments}
In this section, we evaluate the recommendation quality of FedPOIRec on five real-world datasets and we experimentally show that the privacy preserving social features integration protocol incurs low communication and computation cost along with low decryption error. In Section \ref{experimental_setup}, we introduce the datasets, the evaluation metrics and the collaborative filtering algorithms. Then, Section \ref{performance_baseline} compares the recommendation quality of FedPOIRec against the equivalent centralized approaches. Section \ref{performance_social} shows the recommendation enhancement by integrating the social features. Finally, Section \ref{performance_fhe} discusses the communication and computation cost that is introduced by integrating social influence over encrypted data.

We implement the learning process in both centralized and federated setting using PyTorch \cite{paszke_pytorch:_2019}. The privacy preserving social features integration is implemented in PALISADE v1.11.4 \cite{palisade_2021}, which is an open-source homomorphic encryption library. The experiments were conducted on a desktop with AMD Ryzen 9 CPU with 12-cores at 3.8 GHz and GPU support for faster training time on the collaborative filtering algorithms. The security level in CKKS is set to 128 bits according to the recommendation of the homomorphic encryption standard \cite{albrecht_homomorphic_2019}.

\subsection{Experimental Setup}
\label{experimental_setup}
\paragraph{\textbf{Datasets}}

We employ the publicly available dataset from Foursquare collected in \cite{yang_revisiting_2019}, which contains user check-ins and social relationships (friendships) from Twitter. The dataset contains 22,809,624 global-scale check-ins by 114,324 users on 3,820,891 POIs with 363,704 friendships. Without loss of generality, we select five urban areas for top-$N$ POI predictions: Athens, New York, Sao Paolo, Tokyo and Istanbul. 

For dataset preprocessing, we treat the presence of a check-in as a positive interaction. Following the common practice in the literature for POI recommendations \cite{yuan_time-aware_2013, yang_bridging_2017}, we remove users and venues having less than five interactions as cold-start recommendation is treated as a different problem. We then keep users from the largest connected component of the social network for each urban area, following \cite{yang_revisiting_2019, yang_lbsn2vec++:_2020}, to evaluate the social impact on recommendations. After that, we group actions by users and sort the interactions by timestamps. We remove duplicate check-ins and keep the last interaction in a venue for each user as recent actions tend to contain more information for future recommendations \cite{tang_personalized_2018}.

To evaluate the recommendation models, we hold the last 20\% of interactions in each user's check-in data for testing and utilize the remaining 80\% as training data. The statistics of the preprocessed training datasets are summarized in Table \ref{tab:datasetstatistics}. Note that the coverage entry shows the average fraction of interacted POIs that are also visited by the friends of a user.

\begin{table*}[hbt!]
    \centering
    \begin{tabular}{ llllllll }
        \textbf{Dataset} &   \textbf{\#users} & \textbf{\#venues} & \textbf{\#actions} & \textbf{Sparsity}   & \textbf{\#Friendships}    &   \textbf{\#Avg. Friends}    &   \textbf{Coverage}  \\
        \midrule
        Athens   &   310   &   735    &   5,928  & 97.38\% & 1,033  & 6.66452  & 0.31619\\
        New York & 3,516 & 10,107 & 94,654 & 99.73\% & 7,428    & 4.22469  &   0.16498\\
        Sao Paolo & 3,974 & 10,308 & 117,862 & 99.71\% & 9,668   & 4.86563 &   0.24667\\
        Tokyo & 7,241 & 32,515 & 466,787 & 99.80\% & 38,021  &   10.50104    &   0.23856\\
        Istanbul & 10,231 & 23,791 & 522,740 & 99.79\% & 21,231  &   4.15033   &   0.20674\\
        \midrule
  \end{tabular}
  \caption{Statistics of the training datasets.}
  \label{tab:datasetstatistics}
\end{table*}

\paragraph{\textbf{Evaluation Metrics}}
The recommendation quality is measured for each user in the test dataset. Let $Rank$ be the test set for a user $u \in \mathcal{U}$ and $\hat{Rank}$ the predicted ranked list, which contains all POIs that are not part of the training dataset for a user $u$. The predicted ranked list is evaluated by exploiting Precision@K, Recall@K and Mean Average Precision (MAP). In this work, we report precision and recall with $K=\{1,5,10\}$. We also report the F1 score, which is the harmonic mean between precision and recall.

Precision  measures the recommendation quality by identifying the number of recommended venues that are included in $Rank$ against the number of recommended venues by $P@K = Rank \cap \hat{Rank}[:K] \: / \: K$.

Recall is calculated by the number of recommended venues that are included in $Rank$ against the number of venues in the test set by $R@K = Rank \cap \hat{Rank}[:K] \: / \: |Rank|$.

MAP measures the average precision (AP) by exploiting the $P@K$ against the number of venues that are not part of the training dataset for a user by $AP = \sum_{K=1}^{|D_u^-|} P@K \cdot rel(K) \: / \: |D_u^-|$, where $rel(K) = \{0, 1\}$, indicates whether the $K$ venue is included in $Rank$. 

\paragraph{\textbf{Implementation Details}}
To verify the effectiveness of FedPOIRec, we compare it with random recommendations without any learning procedure and the corresponding centralized approaches. In more detail, we consider the following models:
\begin{itemize}
    \item \textbf{Null:} It is the simplest baseline that generates random recommendations (Null-RR) or rank venues according to random embeddings (Null-RE).
    \item \textbf{BPR-MF:} It is a state-of-the-art optimization method based for unary feedback \cite{rendle_bpr:_2009}, which uses a pair of an observed and an unobserved interaction in the training stage and is integrated into matrix factorization. In the training stage, the model learns to provide higher ranks to observed instances by minimizing the following objective function:
        \begin{equation*}
        \label{bpr_loss}
            \mathcal{L}_{BPR} = - \sum_{(u,i,j) \in D_u}ln\sigma(\hat{x}_{uij})
        \end{equation*}
where $\sigma$ is the logistic function and $\hat{x}_{uij} = \hat{x}_{ui} - \hat{x}_{uj}$ with $\{i,j\}$ a pair of observed and unobserved interaction, respectively.
    \item \textbf{CASER:} It is a convolutional neural network that formulates sequences of interactions in chronological order \cite{tang_personalized_2018}. Given the sequence of interactions for a user, the objective function captures the probability that an item is the next in the pattern by minimizing the binary cross-entropy loss:
    \begin{equation*}
    \label{caser_loss}
        \mathcal{L}_{CASER} = - \sum_{s \in D_{u}}\sum_{p \in P} log\sigma(\hat{x}_{u, p}) + \sum_{q \neq p}log(1 - \sigma(\hat{x}_{u, q}))
    \end{equation*}
where $s$ denotes a sequence of interactions, $p \in P$ the next interacted item in the training sequence and $q$ an unobserved sequence of interactions.
    \item \textbf{FedBPR-MF:} It is the BPR-MF algorithm adapted to the federated setting. A similar setting is also presented in \cite{ferrara_fpl_2021}.
    \item \textbf{FedCASER:} It is the CASER sequential approach adapted to the federated setting.
\end{itemize}

The two considered models, i.e., BPR-MF and CASER, can be easily adjusted in the generic FedPOIRec system (Algorithm \ref{fedpoirec_alg}) by modifying the local dataset representation and replacing the objective function. Note that both models formulate unobserved instances to minimize their corresponding loss functions.

For common hyperparameters, we consider the dimension of embeddings $d$ from \{$4, 8, 16, 32, 50, 64, 100, 128$\} and the $l_2$ regularizer from \{$1, 10^{-1}, ..., 10^{-8}$\}. For faster convergence in the centralized setting, we consider the Adaptive Moment Estimation (Adam) \cite{kingma_adam:_2015} optimizer with the learning rate $\eta$ from \{$10^{-1}, 10^{-2}, 10^{-3},$ $10^{-4}$\}. For the federated setting, we consider the SGD optimization method with the learning rate $\eta$ from \{$10^{-1}$, $10^{-2}, 10^{-3}$\}. For other hyperparameters in the CASER model, e.g., the number of convolutional filters, we follow the observations from \cite{tang_personalized_2018}.

After grid search, we fixed the models with the hyperparameters that achieved the highest recommendation quality with respect to three considered metrics. The latent dimension $d$ is fixed to $128$, the learning rate $\eta$ to $10^{-3}$ for the centralized approaches and $10^{-1}$ for the federated models and the $l_2$ regularizer to $10^{-6}$. 

We conduct $150$ training rounds for both the centralized and federated setting and we sample one unobserved interaction for each visit and triplets of random sequences for each training sequence in BPR and CASER, respectively. The learning algorithms are trained from scratch by repeating the experiments 10 times using different initialization seeds and we report the averaged results. In federated variants, we select the 10\% of the users in each global round, following \cite{mcmahan_communication-efficient_2017}. The subset of users that participate in each round is selected randomly such that all groups of participants are equal probable and each participant conducts 5 epochs of local training. At the end of training, we keep the model's parameters at the iteration that achieved the highest quality with respect to the considered metrics.

\subsection{Performance Comparison}
\label{performance_baseline}
\begin{figure*}[htb!]
     \centering
     \includegraphics[width=\textwidth,height=\textheight,keepaspectratio]{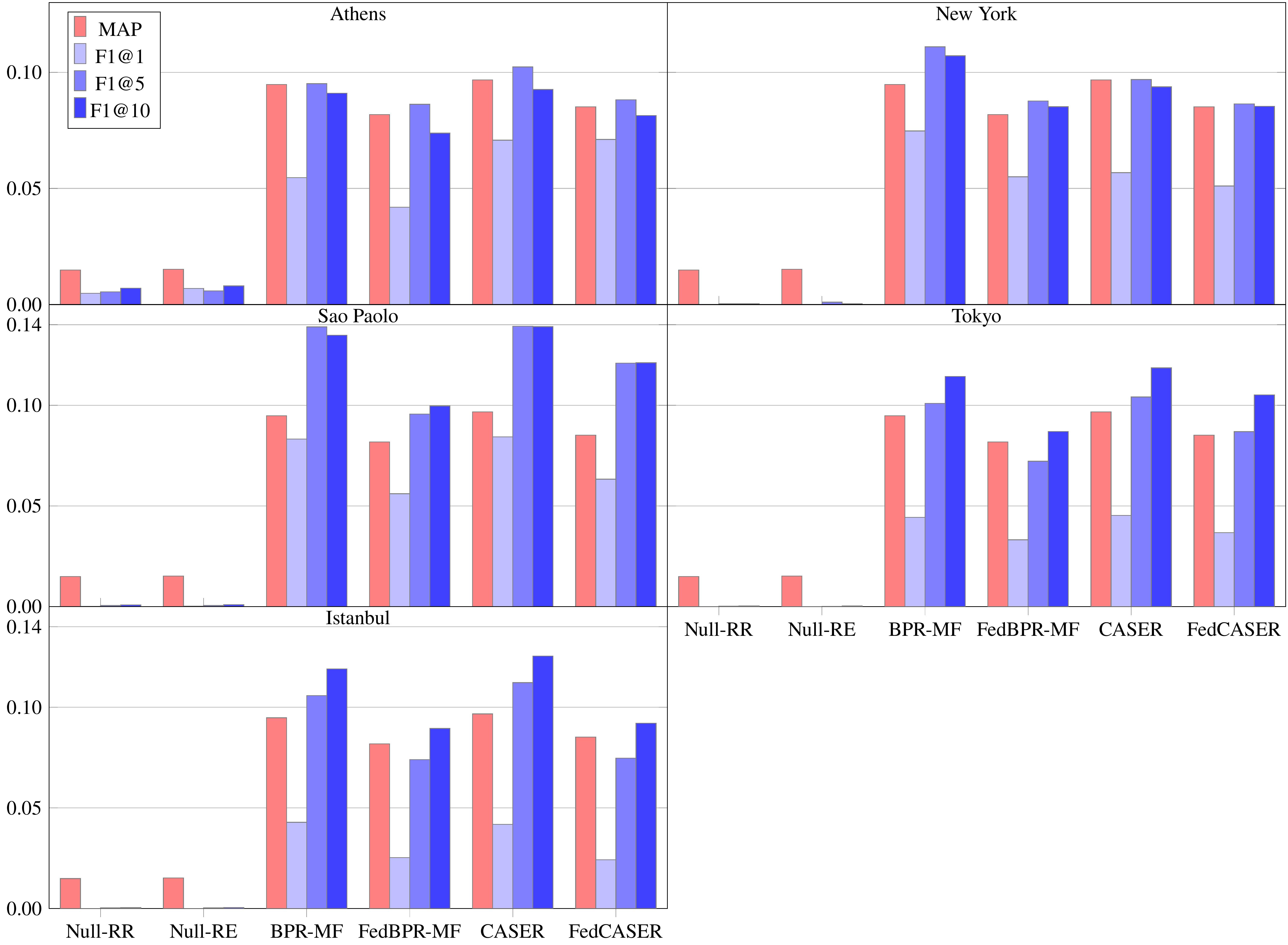}
     \caption{Recommendation quality comparison on the considered models without social features.}
     \label{fig:performance_comparison}
\end{figure*}

\paragraph{\textbf{Quality Comparison}}
The aim of this experimental comparison is to evaluate whether the federated models can achieve comparable recommendation quality to their corresponding centralized approaches as well as to assess whether they outperform random models. Due to space limitations, we present the averaged results for each dataset using MAP and F1 score at $\{1, 5, 10\}$ (Figure \ref{fig:performance_comparison}). The complete results for MAP, precision, recall and F1 score are given in \ref{appendix_a}, Table \ref{tab:rec_comparison}. Note that there is no training phase in Null models. 

From the experimental results, it is observed that the collaborative filtering algorithms in both centralized and federated learning significantly outperform the Null models. To verify the significance of the results we have assessed Student's paired test and we observed p-values close to zero (e.g., $10^{-10}$).

Compared to centralized approaches, the federated models tend to achieve lower recommendation quality. Based on the experimental results, we observe that the bigger the dataset with respect to the number of users and venues, the greater the reduction in recommendation quality. For instance, for the smaller dataset (Athens), the federated models achieve almost equivalent quality with the centralized approaches, i.e., 0.09475 vs 0.08181 and 0.09671 vs 0.08517 in BPR and CASER for the MAP metric, respectively. On the other hand, the recommendation quality of the federated approaches on the dataset with the greater number of users (Istanbul) is almost half the corresponding centralized models, i.e., 0.08976 vs 0.05633 and 0.09219 vs 0.05361, respectively. 

In any case, the federated setting significantly outperforms the Null models. Although the quality decrease compared to the centralized setting, federated models can provide satisfactory recommendations, while offering high privacy guarantees, thus optimizing the privacy-utility trade-off. A way to further enhance the recommendation quality, taking into account the observation that the quality may increase with the drop of the number of users that participate in a global model computation, is to employ a geographical clustering approach to create multiple regional models. We plan to explore such an approach in the future.

\paragraph{\textbf{Efficiency Comparison}}
In the general case, the CASER sequential model provides higher recommendation quality. Although, CASER outperforms the simple BPR model, the communication cost may be prohibitive. In our setting, each client needs to download the global model, operate on it using the local data and then forward the updates to the parameter server. The BPR approach is a simple matrix factorization, i.e., it requires the distribution of a $d \cdot |N|$ matrix, while CASER includes additional convolutional filters and fully connected layers. In particular, the communication cost increases linearly with the number of POIs in the profile. On the larger dataset with respect to the number of venues, i.e., Tokyo which contains 32,515 POIs, the BPR and CASER models, require the transmission of 15 MB and 50 MB of information, respectively. Note that the communication cost can be reduced by adopting a compression strategy, such as the Sparse Ternary Compression method \cite{sattler_robust_2020}. Hence, the adoption of complex models in the federated setting is not trivial as the communication cost may be prohibitive. On a higher level, both models can provide high quality recommendations and therefore, in a real world scenario, a simple matrix factorization is preferable due to its simplicity.

\subsection{Performance Comparison with Social Features Integration}
\label{performance_social}
The decrease in the quality of the federated models is attributed to data heterogeneity among clients, which results in inconsistent local objectives with the global minimum \cite{zhao_federated_2018}. Since the selection of clients in each training round is performed uniformly at random, a finalized global model may not be representative for each user's distribution. Consequently, additional training iterations on the generalized model, minimize local objectives and can lead to quality enhancement.  In addition, collaborative filtering algorithms try to identify patterns between users, which cannot be directly achieved in the federated setting, as local data never leaves the owner's device. The former case can be mitigated using different optimization techniques, e.g., \cite{karimireddy_2020_scaffold}, and personalization steps, while for the latter, we argue that fusing learned parameters between clients, can lead to higher quality recommendations.

\begin{figure*}[htb!]
     \centering
     \begin{subfigure}[b]{0.49\textwidth}
         \centering
         \includegraphics[width=\textwidth,height=0.35\textheight]{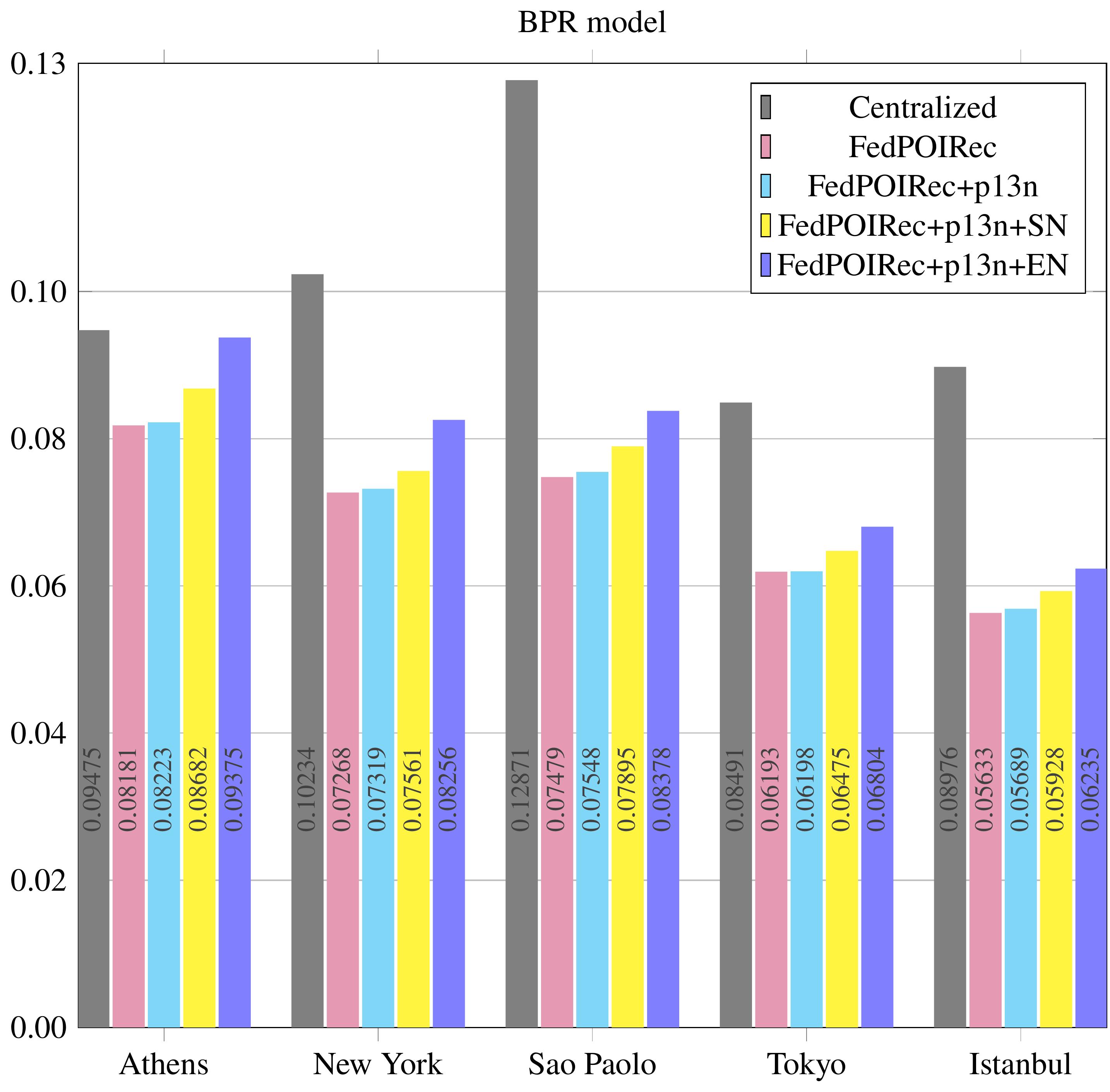}
         \caption{Social influence on FedBPR-MF.}
         \label{fig:BPR_social}
     \end{subfigure}
     \hfill
     \begin{subfigure}[b]{0.49\textwidth}
         \centering
         \includegraphics[width=\textwidth,height=0.35\textheight]{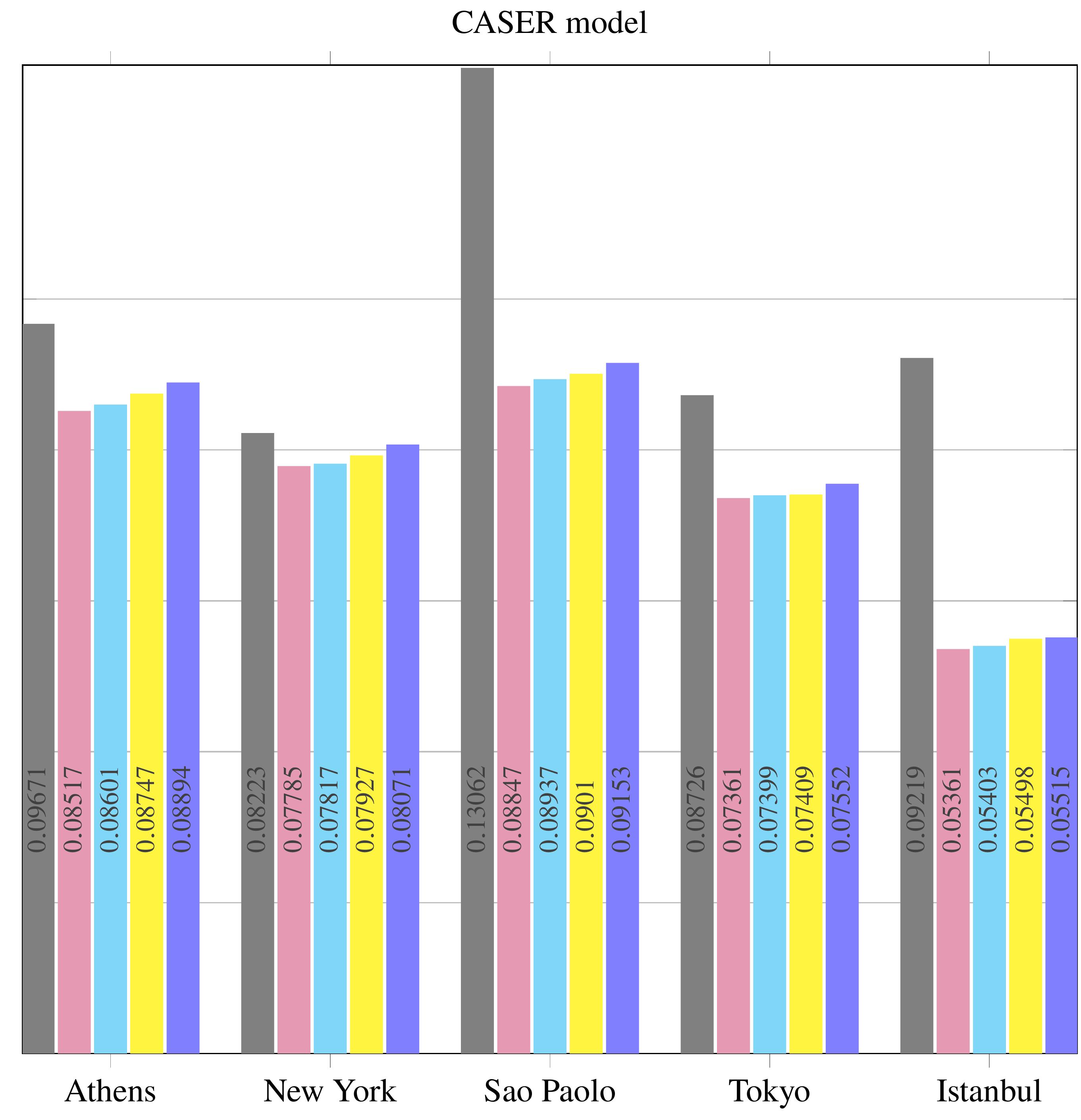}
         \caption{Social influence on FedCASER.}
         \label{fig:CASER_social}
     \end{subfigure}
     \hfill
     \caption{MAP in traditional machine learning, plain FedPOIRec, FedPOIRec enhanced with additional local epochs (FedPOIRec+p13n), FedPOIRec with additional local epochs and social influence (FedPOIRec+p13n+SN) and FedPOIRec with additional local epochs and influence from the entire network of users (FedPOIRec+p13n+EN).}
     \label{fig:bpr_caser_social}
\end{figure*}

Figure \ref{fig:bpr_caser_social} reports the averaged MAP in five settings for the two considered models: i) the centralized approach, which is used as a baseline ii) FedPOIRec with global model inference, iii) FedPOIRec with personalization, i.e., each client locally performs additional training iterations to the finalized global models, iv) FedPOIRec with personalization and social influence, i.e., fusion of learned parameters among 1-hop friends and v) FedPOIRec with personalization and network influence, i.e., fusion of learned parameters among all participating users. 

Compared to the standard federated learning, i.e., global model training, locally trained models after the generation of the global model, slightly increase the recommendation quality in both considered algorithms (Figures \ref{fig:BPR_social} and \ref{fig:CASER_social}). In the standard setting, the model is iteratively generated by considering the intermediate local updates, which are computed based on local observations. 

In the real world, the preferences of friends directly influence an individual's actions. In our setting, the preferences of each user to POIs is reflected in terms of local learned user vectors. From Table \ref{tab:datasetstatistics}, it is evident that on average, the $25\%$ of the users' interacted venues are also visited by their friends. The strongest correlation between a user's visits and direct neighbors' visits in terms of coverage is observed in Athens, while the weakest correlation in New York. Consequently, by fusing the learned parameters, we observed similar behavior in the recommendation quality: in the general case, the quality increases, while the strongest influence is observed in Athens. Going further, with the fusion of the parameters with every user in the dataset, the recommendation quality further increases. However, the increase on the latter case has high computational cost as every local vector needs to be included in the computation. Hence, friendships can approximate strong correlations between users and therefore social information can be utilized to enhance the recommendation quality. In addition, the difference in the quality increase for the five datasets, after the integration of social features, can be attributed to the homogeneity of links in the network as well as to specific characteristics of the population. Hence, the extraction of preference indicators among users from the social network needs to be further investigated.

The introduction of social features enhances the observation that the federated models tend to achieve almost equivalent recommendation quality to the corresponding centralized approaches (Section \ref{performance_baseline}), at least for small datasets regarding the fraction of users - venues. For instance, in Athens, the quality loss of the federated models compared to the corresponding centralized approaches, after integrating the social information, is less than 0.01 in terms of MAP (0.00793 and 0.00924, respectively). Accordingly, the investigation of new aggregation algorithms during the training stage as well as personalization methods after the global model generation, in combination with the integration of social features, may lead the federated setting to generate equivalent recommendations with the conventional machine learning. The combination of the above methods in federated collaborative filtering remains an open issue and we intend to explore it in future work.

Comparing the two considered collaborative filtering algorithms in terms of quality increase by introducing social features integration, it is observed that quality in the BPR case is almost equivalent the corresponding CASER model. This is because in the former case, the predicted ranked list for each user is generated by directly considering the learned latent user vectors, i.e., a dot product calculation between the user's vector and the venues vectors. On the other hand, in the CASER case, the latent user vectors are fed to convolutional filters which are further transformed using fully connected layers, thus decreasing their influence to the generation of recommendations. Considering the fact that CASER also introduces further communication cost due to its complex architecture, BPR or simple matrix factorization techniques may be more effective, at least in the federated setting for collaborative filtering.

Finally, we emphasize that the social features integration should be performed using a privacy preserving approach. The local vectors directly expose the preferences of a particular user and therefore, a passive entity may deduce the local data, thus violating the whole privacy concept. In the next section, we evaluate our privacy preserving computation solution which is based on the CKKS FHE scheme.

\subsection{Privacy Preserving Mean Vector Computation Performance}
\label{performance_fhe}
In this section, we provide the computation and communication cost that is introduced by leveraging our privacy preserving computation protocol for a weighted mean vector generation. Our results show that our protocol introduces negligible cost on the user's side with low decryption error.

On a higher level of our scheme, after receiving the encrypted vectors, a server (cryptoEvaluator) performs computations on encrypted data. The first operation is to calculate the cosine similarity, which requires the computation of the dot product between the querier's and a friend's vector. Recall that we modified the computation of the cosine measure to avoid the division operation. Instead of division, users distribute the inverse of the euclidean norm of their vectors and cryptoEvalutor performs multiplication between the computed dot product and the corresponding euclidean norms. Hence, the operation for calculating the cosine similarity consumes a depth of 2 multiplications. Then, rotation operations are introduced to clone the first slot of the SIMD vector that contains the encrypted result of the cosine similarity to every other slot, depending on the dimension $d$ of the latent factors. Finally, the resulting ciphertext after rotations is multiplied with the friend's vectors, thus consuming another level of multiplication depth. Consequently, our protocol requires a multiplication depth of 3 to correctly perform the introduced computations and output the desired mean vector. Note that the computation overhead and storage requirements increase with the number of the multiplication depth.

For other common parameters on HE schemes, we set the plaintext modulus to 50, which determines the precision of the computations\footnote{The CKKS encoding removes a number of the least significant bits and introduces some noise.}, the ciphertext modulus to 160, which depends on the multiplication depth and the plaintext modulus \cite{takagi_ckks_2017}, and the ciphertext dimension to 16384 to achieve 128-bits security. The ciphertext dimension determines the number of real values that can be encoded into a single vector, which is half the corresponding dimension.

Since CKKS is not an exact arithmetic scheme, the result of a computation approximates the equivalent plaintext calculation. Our results showed that the introduced decryption error is imperceptible, as we observed a maximum error of $3.5\cdot 10^{(-10)}$ and $5.4\cdot 10^{(-11)}$ for the sum of the weighted vectors and the sum of weights compared to their plaintext counterparts, respectively. Therefore, the FHE scheme provides equivalent results with the corresponding plaintext computation, while preserving the privacy of the participants. Hence, we report the computation overhead that it is introduced to each participating entity in a mean vector computation. Table \ref{tab:fhe_running_time_storage} shows the storage requirements and average computation time needed per entity, sorted by operation, beginning with the context generation. 

\begin{table*}[hbt!]
    \centering
    \begin{tabular}{ lrlr }
        \textbf{Operation}    &   \textbf{Time(ms)} & \textbf{Entity} & \textbf{Storage}\\
        \midrule
        ContextGen   &   20.43191   &   cryptoEvaluator    &   2.7 KB  \\
        KeyGen & 15.85362 & User & 1.63 MB \\
        EvalKeyGen & 85.73398 & User & 28.4 MB \\
        Encrypt & 15.76220 & User & 2.2 MB \\
        Ecos (equation \ref{enc_cosine}) & 77.10341 & cryptoEvaluator & 791.9 KB \\
        Weighted Vector (algorithm \ref{eweighted_alg}) & 799.13845 & cryptoEvaluator & 529.6 KB\\
        EvalAdd & 0.97342 & cryptoEvaluator & 529.6 KB\\
        Decrypt & 13.87446 & User & 1.3 KB\\
        \midrule
  \end{tabular}
  \caption{Computation time and storage requirements per operation in a mean vector computation plan considering a single friendship,  with embedding size $d=128$.}
  \label{tab:fhe_running_time_storage}
\end{table*}

\paragraph{\textbf{Storage Requirements and Communication Cost}}
We first focus on the user's side, where the operations involve the public-private key pair and evaluation keys generation, the encryption process and finally the decryption operation. Note that when a user only participates as a friend in a computation, the operations only involve the encryption process. For a computation plan generation, a querier needs 32.23 MB of additional storage for caching and transmitting the required parameters. It is easily observed that the evaluation keys consume most of the storage capacity required for a computation, i.e., 28.4 MB, in total. More precisely, the multiplication and rotation keys consume 3.2 MB of additional storage, respectively, while the sum key needed for a dot product computation consume 22.0 MB. This is because a dot product operation performs a multiplication between two encrypted vectors and then rotations are introduced to calculate the sum of the resulting ciphertext. The public key that needs to be outsourced to each entity that is involved in a computation plan requires 1.1 MB and finally, the encrypted parameters that are transmitted to cryptoEvaluator, i.e., the local vector and the inverse of the euclidean norm require 1.1 MB, respectively. As a final step in the computation, cryptoEvaluator transmits the sum of the weighted vectors and the sum of weights to the querier and therefore 1.1 MB of additional storage is introduced. Note that queriers only need a single transmission of their keys and encrypted data and therefore, heavy communication overhead is avoided. In addition, for the users that only participate as friends in a computation, only the public key of the corresponding querier should be cached and 2.2 MB of information should be shared with cryptoEvaluator. 

\begin{figure*}[htb!]
     \centering
     \begin{subfigure}[b]{0.49\textwidth}
         \centering
         \includegraphics[width=\textwidth,height=0.35\textheight]{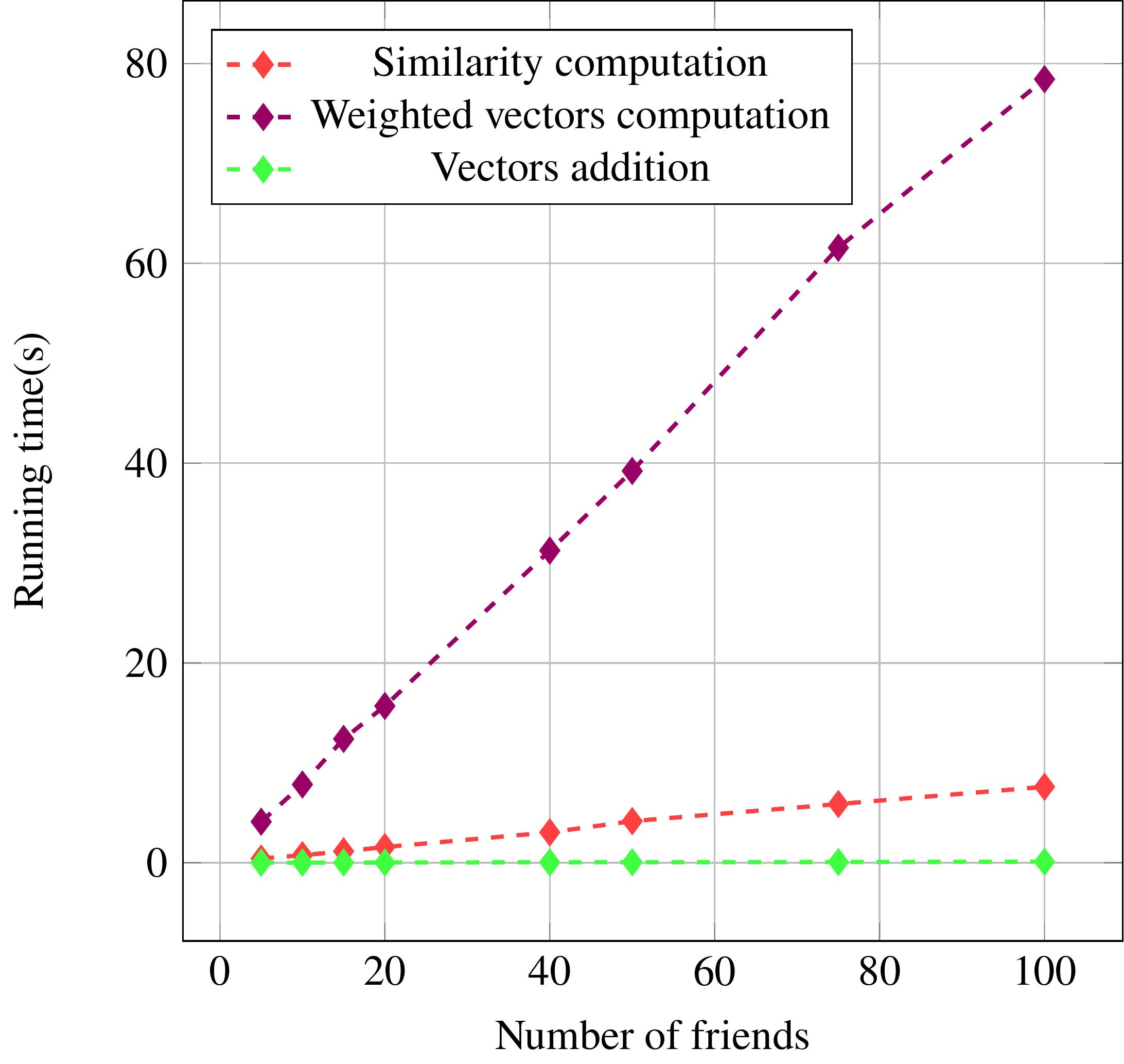}
         \caption{Running time.}
         \label{fig:fhe_running_time_per_num_friends}
     \end{subfigure}
     \hfill
     \begin{subfigure}[b]{0.49\textwidth}
         \centering
         \includegraphics[width=\textwidth,height=0.35\textheight]{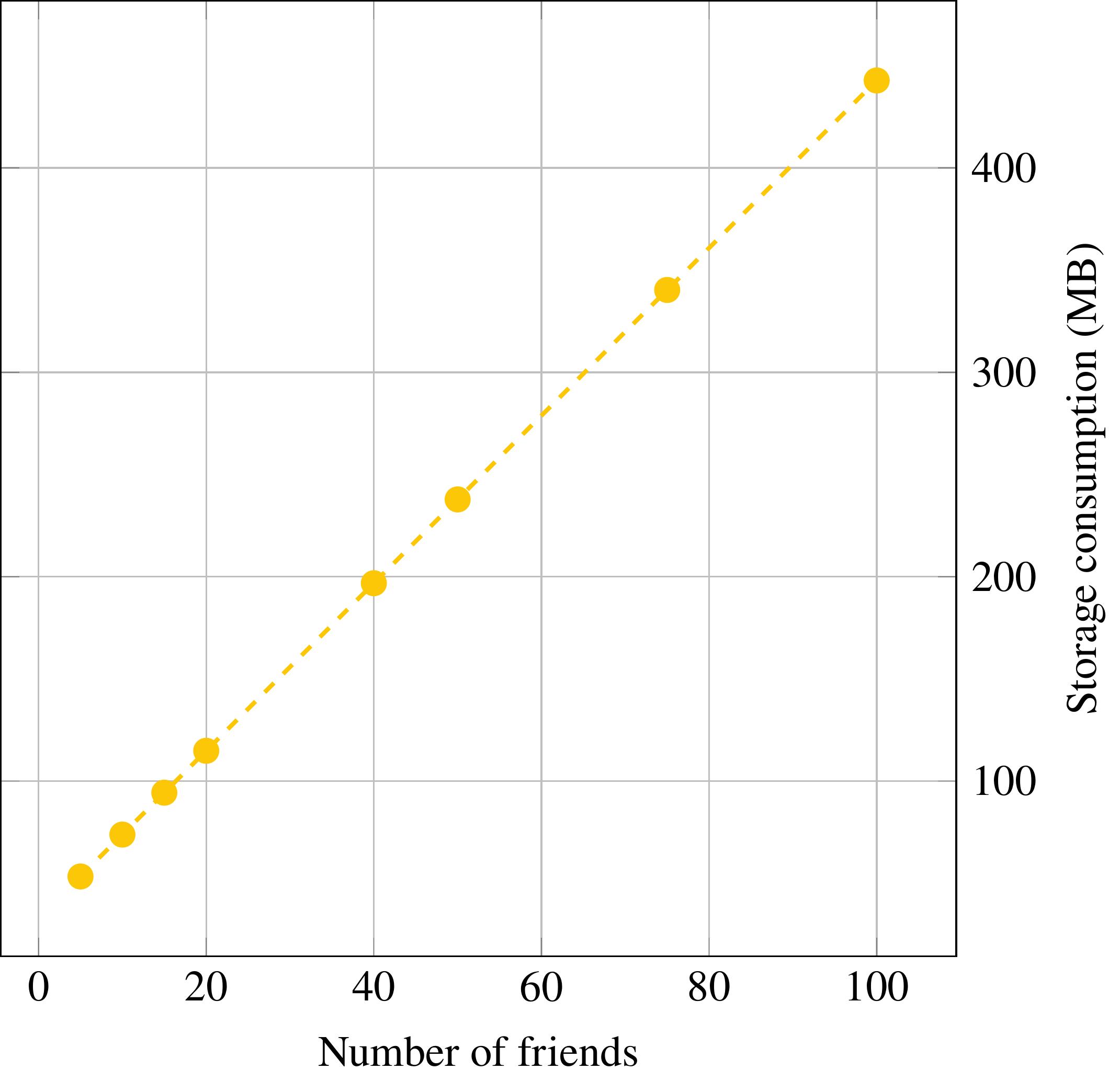}
         \caption{Storage requirements.}
         \label{fig:fhe_storage_per_num_friends}
     \end{subfigure}
     \hfill
     \caption{Running time and storage requirements for a single mean vector computation per $\{5, 10, 515, 20, 40, 50, 75, 100\}$ friends on cryptoEvaluator's side with embedding size $d=128$.}
     \label{fig:fhe_runtime_storage_mean_vector}
\end{figure*}

The mean vector computation operations are performed on the cryptoEvaluator's side. Every required data, i.e., public and evaluation keys as well as encrypted vectors should be cached. Hence, cryptoEvaluator should store the parameters received from a querier as well as the data received from the friends of the querier. Clearly, the storage requirements linearly increase by the number of users who participate in a mean vector computation. Figure \ref{fig:fhe_storage_per_num_friends} shows the increase in storage requirements per number of friends for caching the required keys and the encrypted vectors transmitted from users. Note that additional storage is also required for the intermediate computations, i.e., each resulting cosine similarity and weighted vector require 791.9 KB, respectively. As a final step, the cached weighted vectors and cosine similarities are added together and produce 1.1 MB of data, in total. Although storage requirements are high for caching every outsourced data and intermediate results, cryptoEvaluator is a powerful server that can provide adequate storage.

\paragraph{\textbf{Computation Cost}}
Each operation in our protocol is performed in terms of milliseconds (ms). From Table \ref{tab:fhe_running_time_storage}, it is observed that all users' operations can be executed under a single second. Note however, that our evaluation is performed on a desktop much powerful than devices with limited resources. On the other hand, the computation time that is introduced can be almost negligible considering the advances in CPU constructions of mobile devices. Similar to storage requirements, the heaviest task is the generation of evaluation keys on the querier's side, which requires 85.74 ms of execution. The operations for generating the public-secret key pair as well as the encryption and decryption processes require as little as 16 ms of execution, respectively, and hence, our protocol remains practical on the user's side.

Different from the operations on the users side, at cryptoEvaluator, the execution time is dependent with the number of clients that participate in a computation plan as well as the number of values packed in a single encrypted vector, i.e., the latent dimension in our case. The first operation requires the computation of the cosine similarities between the querier's and the users vectors that responded to the computation plan. In a similar manner, Algorithm \ref{eweighted_alg} for calculating weighted vectors should be executed for each received data. The final step involves addition operations between the resulting ciphertexts. Figure \ref{fig:fhe_running_time_per_num_friends} reports the increase on execution time (in seconds) for a mean vector computation on the server's side per number of users involved in a computation. Again, the heaviest task is the computation of a weighted mean vector, which involves as many rotations as the number of values packed in the encrypted vectors. A single rotation is almost as expensive as multiplication and therefore, a weighted vector computation is the main bottleneck of our approach. On the other hand, even in extreme cases, i.e., when many users respond to a computation plan, the whole process can be executed within a few minutes. For instance, the execution time for a computation plan with 100 users requires about 90 seconds of operations with 128 values packed into a single ciphertext and hence, the protocol remains efficient. 
 
 Summarizing, high storage requirements and communication costs are transferred to a server. A querier only needs a single transfer of the required data for a computation to cryptoEvaluator and after calculations only two encrypted vectors are transmitted back to the querier for decryption, i.e., 2 communication rounds are needed. In a similar manner, the users that participate in a computation, receive the corresponding public key and only transmit their encrypted vectors to cryptoEvaluator, i.e., 2 communication rounds are required. Hence, only 2 communication rounds are introduced on each user's side and thus, heavy communication costs are avoided. In a similar manner, the heaviest computational tasks (similarity and mean vector computations) are transferred to the server's side and therefore, additional overhead is not introduced on users' devices. Figure \ref{fig:fhe_runtime_storage_mean_vector} summarizes the computation overhead and the storage requirements that are introduced at cryptoEvaluator per number of users that participate in a computation plan. Even when the number of users that participate in a computation plan is high, a complete run can be executed fairly quick and the protocol remains efficient. Hence, our privacy preserving protocol is practical, communication efficient as well as it does not require heavy storage requirements on the users' side.
 
\section{Conclusion}
\label{conclusion}
In this paper, we presented FedPOIRec, a generic practical privacy preserving solution for POI recommendations enhanced with knowledge transfer between friends by operating on encrypted data. The federated learning procedure complies with data dissemination and minimization regulations, as local data is maintained in the owner's device. The intermediate computations are summarized using a secure aggregation strategy based on secure multiparty computation to prevent passive entities from deducing additional information. The social features integration incurs after the federated process, where a server operates over encrypted data and fuses learned user vectors using the properties of a fully homomorphic encryption scheme. Our privacy preserving protocol for social fusion can also be adapted to other cases and tasks, when a privacy preserving computation based on local data is required.

The experimental results demonstrate that although federated learning leads to quality decrease compared to traditional machine learning approaches, it is a promising solution with high privacy guarantees. The privacy preserving social features integration protocol enhances the recommendation quality, while incurring low communication and computation overhead on the users' side, as it requires only two rounds of communication between a querier and the server, with low decryption error.

Possible future directions are twofold: federated learning and privacy preserving based. For the federated setting, further analysis and sophisticated algorithms on personalization should be investigated, evaluation of the optimization algorithms which can lead to the generation of more generalized and high quality models should be performed, and ways for formulating social features integration during the training stage with communication efficiency should be proposed. Regarding privacy, it is crucial to carry out a formal security analysis on the security and privacy guarantees offered by the existing secure aggregation schemes, the adaptation of the proposed privacy preserving protocol for a weighted mean vector computation in other learning tasks, such as clustering, should be evaluated as well as ways for further reduction on the computation and communication overhead using fully homomorphic schemes should be explored.

\section*{Acknowledgements}
This work has been co-financed by the European Union and Greek national funds through the Operational Program Competitiveness, Entrepreneurship and Innovation, under the call RESEARCH -- CREATE -- INNOVATE (project code: T1EDK-02474, grant no.: MIS 5030446).

 \bibliographystyle{elsarticle-num} 
 \bibliography{cas-refs}
 
\appendix
\setcounter{table}{0}
\section{Complete Experimental Results}
\label{appendix_a}
\begin{table*}[hbt!]
    \scriptsize
    \centering
    \begin{tabular}{ l|l|cccccc }
        \hline
        \textbf{Dataset}    &   \textbf{Metric} & \textbf{Null-RR} & \textbf{Null-RE}  & \textbf{BPR-MF} & \textbf{FedBPR-MF} & \textbf{CASER} & \textbf{FedCASER} \\
        \hline
        \multirow{10}{4em}{Athens} & MAP & 0.01491 & 0.01517 & 0.09475 & 0.08181 & 0.09671  & 0.08517\\ 
        & P@1 & 0.01969 & 0.06432 & 0.16481 & 0.16010 & 0.18394 & 0.18061\\ 
       & P@5 &  0.00483 & 0.00526 & 0.09173 & 0.08365 & 0.10194 & 0.08560\\ 
        & P@10 &  0.00501 & 0.00591 & 0.06781 & 0.05452 & 0.06971 & 0.06142\\
        & R@1 & 0.00278 & 0.00366 & 0.03274 & 0.02411 & 0.04382 & 0.04426\\
        & R@5 &  0.00645 & 0.00679 & 0.09878 & 0.08899 & 0.10273 & 0.09089\\
        & R@10 &  0.01178 & 0.01263 & 0.13814 & 0.11432 & 0.13782 & 0.12049\\ 
        & F1@1 & 0.00487 & 0.00693 & 0.05463 & 0.04191 & 0.07078 & 0.07110\\
        & F1@5 & 0.00552 & 0.00593 & 0.09512 & 0.08624 & 0.10233 & 0.08817\\
        & F1@10 & 0.00703 & 0.00805 & 0.09097 & 0.07383 & 0.09259 & 0.08136\\
        \hline
        \multirow{10}{5em}{New York} & MAP & 0.00141 & 0.00161 & 0.10234  & 0.07268 & 0.08223  & 0.07785 \\ 
        & P@1 & 0.00314 & 0.00037 & 0.24282  & 0.16519 & 0.17824 & 0.16348\\ 
        & P@5 & 0.00029  & 0.00314 & 0.12336 & 0.09755 & 0.10323 & 0.09613\\ 
        & P@10 &  0.00017  & 0.00022 & 0.08407 &  0.06752 & 0.07397 & 0.06934\\
       & R@1 & 0.00001  & 0.00001  & 0.04417 & 0.03302 & 0.03378 & 0.03025\\
        & R@5 & 0.00049 & 0.00062 & 0.10086 & 0.07960 &  0.09124 & 0.07843\\
        & R@10 &  0.00176 & 0.00209 & 0.14759  & 0.11543 & 0.12792 & 0.11093\\
        & F1@1 & 0.00002 & 0.00002 & 0.07474 & 0.05504 & 0.05680 & 0.05105\\
        & F1@5 & 0.00036 & 0.00104 & 0.11098 & 0.08767 & 0.09687 & 0.08638\\
        & F1@10 & 0.00031 & 0.00040 & 0.10712 & 0.08520 & 0.09374 & 0.08534\\
        \hline
        \multirow{10}{5em}{Sao Paolo} & MAP & 0.00138 & 0.00145 & 0.12871 & 0.07479 & 0.13062  & 0.08847\\ 
        & P@1 & 0.00071  & 0.00075 & 0.31776 & 0.20878 & 0.32067 & 0.24791\\ 
        & P@5 & 0.00069 & 0.00066 & 0.15978  & 0.11871 & 0.15986 & 0.14716\\ 
        & P@10 & 0.00057 & 0.00059  & 0.11758 & 0.08873 & 0.12371 & 0.10385\\
        & R@1 &  0.00007  & 0.00009 & 0.04786 & 0.03238 & 0.04851 & 0.03628\\
        & R@5 & 0.00042  & 0.00045 & 0.12291 & 0.08004 & 0.12337 & 0.10264\\
        & R@10 &  0.00135 & 0.00268 & 0.15792 & 0.11365 & 0.15871 & 0.14522\\
        & F1@1 & 0.00013 & 0.00016 & 0.08319 & 0.05606 & 0.08427 & 0.06330\\
        & F1@5 & 0.00052 & 0.00054 & 0.13894 & 0.09561 & 0.13926 & 0.12093 \\
        & F1@10 & 0.00080 & 0.00097 & 0.13480 & 0.09966 & 0.13904 & 0.12110\\
        \hline
        \multirow{10}{5em}{Tokyo} & MAP  & 0.00081 & 0.00086 & 0.08491 & 0.06193 & 0.08726 & 0.07361 \\ 
        & P@1 & 0.00029 & 0.00017 & 0.33271 & 0.26142 & 0.36730 &  0.28609   \\ 
        & P@5 & 0.00034  & 0.00024 & 0.20075 & 0.16051 & 0.21351 &  0.18071\\ 
       & P@10 & 0.00047 & 0.00051 & 0.14492 & 0.12403 &  0.15087 & 0.13207\\
        & R@1 & 0.00001 & 0.00001 & 0.02376 & 0.01771 & 0.02414  & 0.01961\\
        & R@5 & 0.00013 & 0.00011 & 0.06743 & 0.04658 &  0.06884  & 0.05721\\
        & R@10 & 0.00029 & 0.00031  & 0.09431 & 0.06691 &  0.09766 & 0.08737\\
        & F1@1 & 0.00002 & 0.00002 & 0.04435 & 0.03317 & 0.04530 & 0.03670\\
        & F1@5 & 0.00019 & 0.00015 & 0.10095 & 0.07221 & 0.10411 & 0.08691 \\
        & F1@10 & 0.00036 & 0.00039 & 0.11426 & 0.08693 & 0.11857 & 0.10517\\
        \hline
        \multirow{10}{5em}{Istanbul} & MAP & 0.00096  & 0.00089 & 0.08976 & 0.05633 & 0.09219 & 0.05361 \\ 
        & P@1 & 0.00039  & 0.00042 & 0.28448 & 0.10622 & 0.30038 & 0.09896\\ 
        & P@5 &  0.00059 & 0.00058 & 0.17346 & 0.11890 & 0.18992 & 0.12082\\ 
        & P@10 & 0.00042  & 0.00046 & 0.13275 & 0.10814 & 0.13635 & 0.11351\\
        & R@1 &  0.00001  & 0.00001  & 0.02317 & 0.01440 & 0.02248 & 0.01379\\
        & R@5 &  0.00023 & 0.00024 & 0.07612 & 0.05370 & 0.07964 & 0.05404\\
        & R@10 & 0.00052  & 0.00054 & 0.10783 & 0.07627 & 0.11606 & 0.07741\\ 
        & F1@1 & 0.00002 & 0.00002 & 0.04285 & 0.02536 & 0.04183 & 0.02421\\
        & F1@5 & 0.00033 & 0.00034 & 0.10581 & 0.07399 & 0.11222 & 0.07468\\
        & F1@10 & 0.00046 & 0.00050 & 0.11900 & 0.08945 & 0.12539 & 0.09205\\
        \hline
  \end{tabular}
  \caption{Averaged results on the considered models without social features integration after repeating the experiments 10 times from scratch.}
  \label{tab:rec_comparison}
\end{table*}
 






\end{document}